\documentclass[fleqn,10pt]{wlscirep}
\usepackage[utf8]{inputenc}
\usepackage[T1]{fontenc}
\usepackage{newtxtext,newtxmath}
\usepackage{setspace}
\usepackage{indentfirst}
\usepackage{graphicx}
\usepackage{multirow}
\usepackage{makecell}
\usepackage{amsmath}
\usepackage{textcomp}

\newcommand{\fulltablenote}[1]{%
\DeclareRobustCommand{\cosenior}{\ensuremath{\dagger}}
\par\smallskip
\noindent\begin{minipage}{\textwidth}
\raggedright #1
\end{minipage}\par
}
\usepackage{longtable}
\usepackage{array}
\usepackage{tabularx}
\usepackage{colortbl}
\usepackage{amsmath}
\usepackage{url}
\usepackage{xcolor}
\usepackage{CJKutf8}
\usepackage[most]{tcolorbox}

\usepackage{bm}
\usepackage{marvosym}

\usepackage[htt]{hyphenat}
\usepackage{fancyvrb}
\usepackage{fvextra} % 支持自动换行
\usepackage{booktabs, threeparttable, adjustbox, caption, float}

\newcommand{\zh}[1]{\begin{CJK}{UTF8}{gbsn}#1\end{CJK}}

\definecolor{bothcorrect}{HTML}{009E73}
\definecolor{gptcorrsecinc}{HTML}{E69F00}
\definecolor{gptincseccorr}{HTML}{0072B2}
\definecolor{bothincorrect}{HTML}{CC79A7}
\definecolor{sugcorrect}{HTML}{2A9D8F}
\definecolor{sugwrong}{HTML}{C44E3A}

\tcbset{casesbox/.style={
  enhanced, breakable,
  colback=black!2, colframe=black!45,
  boxrule=0.6pt, arc=2pt,
  left=8pt, right=8pt, top=8pt, bottom=8pt,
  fonttitle=\bfseries\sffamily,
  coltitle=white, colbacktitle=black!65
}}
\tcbset{modelbox/.style={
  enhanced,
  boxrule=0.5pt, arc=2pt,
  left=6pt, right=6pt, top=5pt, bottom=5pt,
  fonttitle=\bfseries\sffamily,
  coltitle=white
}}
\newcommand{\casemeta}[5]{%
\begin{minipage}[t]{0.30\linewidth}
\vspace{0pt}
\centering
\includegraphics[width=\linewidth]{#1}
\end{minipage}%
\hspace{2em}%
\begin{minipage}[t]{0.58\linewidth}
\vspace{0pt}
\textbf{Case ID:} #2\\[2pt]
\textbf{Reference diagnosis:} #3\\[2pt]
\textbf{Clinical history:} #4\\[2pt]
\textbf{Concordance:} #5
\end{minipage}}

\title{Dual- versus Single-Suggestion AI Support for Radiographic Interpretation in Residents: Randomized Multireader Study}

\author[1,2,*]{Lin Wu}
\author[3,*]{Zhe Xu}
\author[3]{Hongyi Wang}
\author[4]{Feifei Zhou}
\author[5]{Wei Deng}
\author[1,2]{Chunlong Zhang}
\author[1,2]{Yuting Zhu}
\author[1,2]{Kaixiao Chen}
\author[1,2]{Xiao Liang}
\author[1,2]{Chen Yang}
\author[1,2,\Letter]{Yeyuan Chen}
\author[3,6,7,8,9,\Letter]{Hao Chen}
\author[1,2,\Letter]{Fuqing Zhou}

\affil[1]{Department of Radiology, The First Affiliated Hospital, Jiangxi Medical College, Nanchang University}
\affil[2]{Jiangxi Province Medical Imaging Research Institute}
\affil[3]{Department of Computer Science and Engineering, Hong Kong University of Science and Technology, Hong Kong, China}
\affil[4]{Department of Radiology, Shangrao City People’s Hospital, Shangrao, China}
\affil[5]{Department of Radiology, Nanchang People's Hospital, Nanchang, China}

\affil[6]{Department of Chemical and Biological Engineering, Hong Kong University of Science and Technology, Hong Kong SAR, China}
\affil[7]{Division of Life Science, Hong Kong University of Science and Technology, Hong Kong SAR, China}
\affil[8]{State Key Laboratory of Nervous System Disorders, The Hong Kong University of Science and Technology, Hong Kong SAR, China}
\affil[9]{HKUST Shenzhen-Hong Kong Collaborative Innovation Research Institute, The Hong Kong University of Science and Technology, Futian, Shenzhen, China}

\affil[*]{Lin Wu and Zhe Xu contributed equally to this work.}
\affil[\Letter]{Corresponding Authors}
\affil[ ]{\textbf{Contact: Fuqing Zhou (ndyfy02301@ncu.edu.cn)}}
\begin{abstract}

\textbf{Purpose:} To compare dual- and single-suggestion AI support for radiographic interpretation by residents, particularly when the shared AI suggestion was incorrect.

\textbf{Materials and Methods:} This prospective, multicenter, randomized three-arm reader study was conducted at three hospitals in China from July to September 2026 (ChiCTR2600129243). After specialty stratification, 132 residents with fewer than 3 years of clinical experience were randomized 1:1:1 to GPT-5.4 alone (group A), GPT-5.4 plus Kimi-K2.6 (group B), or GPT-5.4 plus Gemini-3.6 Flash (group C); 123 were analyzed. Participants interpreted 60 radiographs before and after AI support. The primary outcome was accuracy change. Welch ANOVA and Holm-adjusted $t$ tests compared support conditions; HC3 linear models assessed specialty interaction.

\textbf{Results:} Among 123 residents (mean age, 24.1 years $\pm$ 1.4; 65 women), radiology residents showed greater accuracy improvement with dual- than single-suggestion support (B–A, 6.69 percentage points [95\% CI, 0.97–12.40]; C–A, 7.87 percentage points [95\% CI, 1.64–14.11]; Holm-adjusted $P$ = .030 for both), whereas accuracy change did not differ in non-radiology residents ($P$ = .20). When GPT-5.4 was incorrect, AI-assisted accuracy was higher with dual- than single-suggestion support in radiology residents (40.1\% and 40.4\% vs 20.0\%) and non-radiology residents (31.3\% and 31.0\% vs 12.1\%) (all Holm-adjusted $P$ < .001). The dual-suggestion effect differed by specialty (interaction difference, 10.44 percentage points; 95\% CI, 4.36–16.52; $P$ < .001).

\textbf{Conclusion:} Dual-suggestion support may mitigate the influence of erroneous AI suggestions, with greater accuracy improvement observed in radiology but not non-radiology residents.

\end{abstract}

\begin{document}
\maketitle

\section{Introduction}

Artificial intelligence (AI) has substantial potential to improve the accuracy and efficiency of radiographic interpretation\cite{gommers2025ai,schramm2026explanations,hanneman2026promise}. By combining imaging findings with clinical information, AI can provide diagnostic suggestions and supporting rationales\cite{duron2025gpt,tanno2025multimodal}. However, the effectiveness of AI assistance depends not only on model performance but also on how clinicians respond to AI suggestions, particularly when those suggestions are erroneous\cite{thrall2024ai,angus2025ai}. 

Previous studies have examined how information presentation within a single-AI framework influences clinicians’ responses to AI suggestions, including diagnostic conclusions alone or with confidence scores\cite{prinster2024explain,tang2023confidence}, content-based image retrieval\cite{xu2026radiology}, visual heatmaps\cite{xu2026radiology}, and text-based explanations\cite{schramm2026explanations,song2026llm}. Evidence suggests that less experienced physicians may derive greater benefit from AI assistance, yet they are also more susceptible to erroneous AI suggestions\cite{dratsch2023automation,prinster2024explain,teng2026overreliance}. Failure to recognize and reject erroneous AI suggestions can lead to diagnostic errors\cite{park2026beyond,taib2026visual}. Limiting the diagnostic influence of erroneous AI suggestions among less experienced physicians is an important objective in human–AI interaction\cite{angus2025ai,vaccaro2024combinations}.

Drawing on multiple independent opinions is an established approach to reducing reliance on a single judgment, consistent with the principle of collective intelligence\cite{riedl2021collective,woolley2024collective}. In medical diagnosis, combining multiple independent judgments has been shown to outperform individual judgments across a range of medical cases\cite{barnett2019collective}. Similarly, structured information sharing among clinicians has been shown to reduce diagnostic errors and improve individual clinical decisions compared with isolated reflection\cite{centola2023networks}. Applied to AI-assisted diagnosis, presenting two independent AI suggestions may provide an additional diagnostic reference when one suggestion is erroneous\cite{zoller2025humana}. However, whether dual-suggestion support can mitigate the diagnostic impact of erroneous AI suggestions among residents during radiographic interpretation remains unclear.

Building on the principle of collective intelligence\cite{zoller2025humana,kameda2022aggregation}, we evaluated dual-suggestion support as a human–AI interaction strategy. We hypothesized that dual-suggestion support would show greater improvement in diagnostic accuracy than single-suggestion support, particularly when the shared AI suggestion was incorrect. We conducted a three-arm randomized study in which participants interpreted each case first without AI assistance and then with the assigned single- or dual-suggestion support. The primary outcome was change in diagnostic accuracy after AI assistance. Diagnostic accuracy was also assessed according to whether the shared AI suggestion was correct or incorrect. Secondary outcomes included interpretation time and confidence change, and effects were also evaluated by specialty. The purpose of this study was to compare dual- and single-suggestion AI support for radiographic interpretation by residents, particularly when the shared AI suggestion was incorrect.

\section{Materials and Methods}

\subsection{Study Design and Participants}
\begin{figure}[H]
\centering
\includegraphics[width=\linewidth]{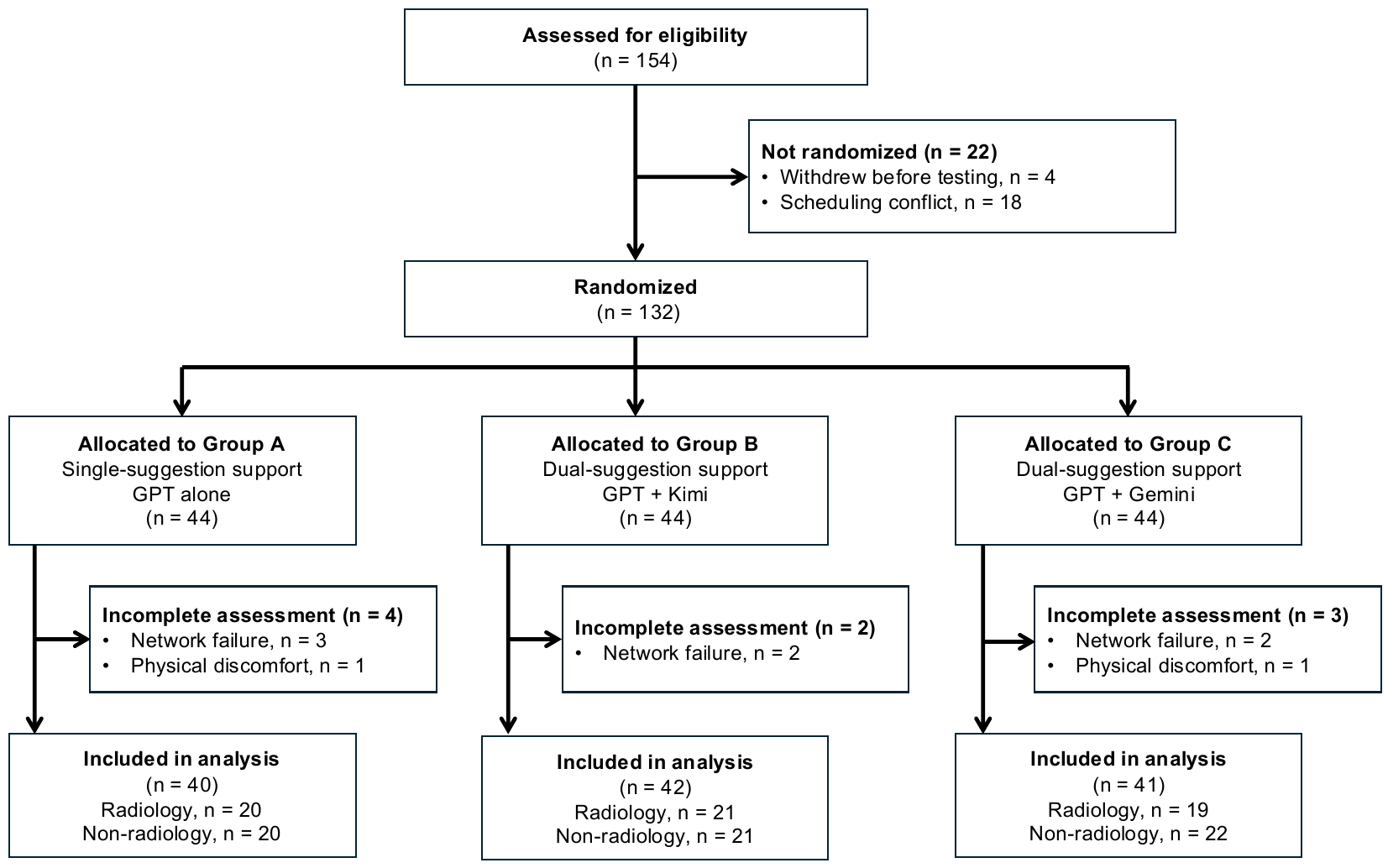}
\caption{\textbf{Participant Flow Diagram.} Participants were stratified by specialty and allocated 1:1:1 to the three AI support conditions. Testing was performed using a smartphone or tablet. Network failure was defined as connectivity or webpage failure preventing image display or response submission; physical discomfort was defined as feeling unwell enough to interfere with concentration and require termination of testing. Final analytic samples are shown by specialty.}
\label{fig:1}
\end{figure}

This multicenter, prospective, randomized, three-arm, parallel-group multireader study was conducted at three hospitals in China from July to September 2026. The study was approved by the institutional ethics committee (2026-756), which waived informed consent, and was registered with the Chinese Clinical Trial Registry (ChiCTR2600129243). Residents were recruited through advertisements. Eligibility required a valid Chinese physician license, prior training in radiographic interpretation, and fewer than 3 years of clinical experience. Participants were classified a priori as radiology residents or non-radiology residents from internal medicine, general surgery, or emergency medicine. Participants with incomplete assessments due to network interruption or physical discomfort were excluded from the final analysis (Figure~\ref{fig:1}).

\subsection{Case Selection and AI Output Generation}
We retrospectively assembled 218 chest and abdominal radiography examinations performed at Center 1 from January 2023 to June 2024 according to prespecified diagnostic categories. Examinations represented four chest and three abdominal conditions, were of sufficient image quality, and had diagnoses confirmed pathologically or clinically. GPT-5.4 (OpenAI), Kimi-K2.6 (Moonshot AI), and Gemini-3.6 Flash (Google) independently generated a primary diagnosis and supporting rationale from the same deidentified radiograph and clinical history\cite{xu2026radiology,song2026llm}. Outputs were generated once, fixed before case selection, and used unchanged throughout the study. To evaluate suggestion number rather than differences in overall model performance, 60 cases were purposively selected to balance disease composition and match overall model accuracy. The final set comprised 30 chest and 30 abdominal radiographs; each model correctly diagnosed 39 of 60 cases (65.0\%), while Kimi-K2.6 and Gemini-3.6 Flash differed in case-level concordance with GPT-5.4. Before the reader study, five radiologists with 1–5 years of experience independently interpreted the 60 cases without AI assistance; mean accuracy was 54.3\%. Additional details are provided in Supplementary Appendix 1.

\subsection{Study Procedure}
\begin{figure}[htbp]
\centering
\includegraphics[width=\linewidth]{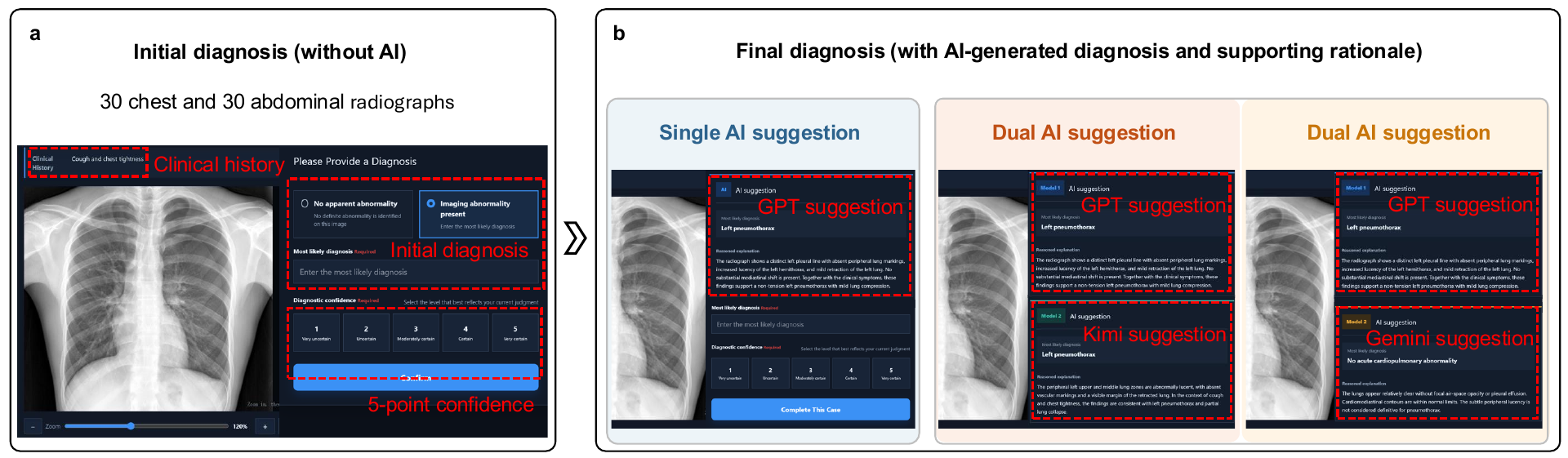}
\caption{\textbf{Reader Workflow and AI Support Conditions.} \textbf{a,} Initial interpretation without AI assistance, including diagnosis and confidence. \textbf{b,} Final interpretation after AI assistance with single-suggestion support (group A, GPT-5.4) or one of two parallel dual-suggestion conditions (group B, GPT-5.4 plus Kimi-K2.6; group C, GPT-5.4 plus Gemini-3.6 Flash). In both dual-suggestion conditions, the two AI suggestions were presented simultaneously.}
\label{fig:2}
\end{figure}

Participants were stratified by specialty and randomly assigned in a 1:1:1 ratio to groups A, B, or C using a prespecified computer-assisted allocation procedure. The allocation list was fixed before assessment and concealed from enrollment personnel until assignment. Details of the allocation procedure are provided in the Supplementary Methods. Group A received GPT-5.4 alone; groups B and C received GPT-5.4 plus Kimi-K2.6 and GPT-5.4 plus Gemini-3.6 Flash, respectively. The two dual-suggestion conditions assessed consistency across model pairings. GPT-5.4 was the shared model; both AI suggestions were presented simultaneously and consisted of a primary diagnosis and supporting rationale. Model identities were not disclosed to participants. All groups interpreted the same 60 cases in independently randomized order.
Testing occurred in supervised group sessions using a smartphone or tablet and was completed in one session. Participants completed the assessment independently using deidentified radiographs and clinical histories. For each case, participants recorded an unaided diagnosis and confidence on a 5-point scale; responses were then locked. The assigned AI suggestion or suggestions were displayed, after which participants recorded a final diagnosis and confidence. The platform recorded responses and interpretation times (Figure~\ref{fig:2}).

\subsection{Outcomes and Response Adjudication}
The primary outcome was change in diagnostic accuracy, defined as AI-assisted minus AI-unassisted accuracy. Accuracy was also evaluated according to whether the shared GPT-5.4 suggestion was correct (39 cases) or incorrect (21 cases). Secondary outcomes were interpretation time, defined as the total time across the unaided and AI-assisted interpretation stages, and confidence change, defined as final minus initial mean confidence. Diagnostic revisions were classified as beneficial (incorrect-to-correct) or harmful (correct-to-incorrect).
Responses were scored using a prespecified case-specific diagnostic lexicon (Supplementary Appendix 2) established by two senior radiologists (>10 years' experience). Free-text responses were classified by an automated matching pipeline and independently reviewed by both radiologists while masked to participant identity and support condition; disagreements were resolved by consensus.

%The primary outcome was change in diagnostic accuracy after AI assistance, defined as AI-assisted minus AI-unassisted accuracy. Diagnostic accuracy was also evaluated separately according to whether the shared GPT-5.4 suggestion was correct (39 cases) or incorrect (21 cases). Secondary outcomes were interpretation time and confidence change. Diagnostic revisions were analyzed as components of accuracy change, comprising beneficial revisions (incorrect-to-correct) and harmful revisions (correct-to-incorrect).

%Participant responses were scored using a prespecified case-specific diagnostic lexicon (Supplementary Appendix 2), which was established by two senior radiologists (L.W.\ and W.D., each with $>$10 years of experience) and included acceptable diagnoses and synonyms. Free-text responses were initially classified by an automated matching pipeline, and all classifications were independently reviewed by both radiologists while masked to participant identity and support condition. Disagreements were resolved by consensus.

\subsection{Statistical Analysis}
Sample size was based on feasible recruitment; no formal a priori calculation was performed. Data are summarized as mean $\pm$ SD or counts and percentages. Paired $t$ tests assessed changes in diagnostic accuracy before and after AI assistance. Between-specialty comparisons used Welch $t$ tests. Prespecified primary comparisons were accuracy changes in groups B and C versus group A within each specialty. Support conditions were compared using Welch ANOVA, with B–A and C–A contrasts assessed using Welch $t$ tests with Holm adjustment. Accuracy before and after AI assistance was also compared across support conditions for shared GPT-5.4-correct and -incorrect cases using the same approach.

Linear models with heteroskedasticity-robust standard errors (HC3 estimator) assessed specialty interactions for accuracy change and beneficial and harmful revisions\cite{mansournia2021robust}. The dual-suggestion effect was defined as  $[(B + C)/2] - A$, and interactions were assessed using Wald tests. Effect estimates are reported with 95\% CIs.  No important changes to the study methods or prespecified outcomes were made after trial commencement. All tests were two-sided, with $P$ < .05 indicating statistical significance. Analyses were performed using Python, version 3.12.14.

%No formal a priori sample size calculation was performed. The sample size was determined pragmatically based on feasible recruitment of eligible residents across the three participating centers during the study period. Data are summarized as mean $\pm$ SD or counts and percentages. Paired $t$ tests assessed changes in diagnostic accuracy before and after AI assistance. The prespecified primary comparisons were changes in diagnostic accuracy in groups B and C versus group A within each specialty stratum. Support conditions were compared using Welch ANOVA, with B--A and C--A comparisons performed using Welch $t$ tests and Holm adjustment.

%Diagnostic accuracy was also compared across support conditions separately for cases in which the shared GPT-5.4 suggestion was correct or incorrect. Linear models with heteroskedasticity-robust standard errors (HC3 estimator) assessed specialty interactions for accuracy change and beneficial and harmful revisions\cite{mansournia2021robust}. The dual-suggestion effect was defined as $[(B + C)/2] - A$, and interactions were assessed using Wald tests. Effect estimates are reported with 95\% CIs. Detailed contrasts are provided in the Supplementary Materials. No important changes to the study methods or prespecified outcomes were made after trial commencement. All tests were two-sided, with $P < .05$ indicating statistical significance. Analyses were performed using Python, version 3.12.14.

\section{Results}

\subsection{Participant Flow and Characteristics}

Of 154 residents assessed for eligibility, 132 were allocated to the three support conditions; nine had incomplete assessments because of network failure (n = 7) or physical discomfort (n = 2), leaving 123 residents for analysis. The 123 residents had a mean age of 24.1 years $\pm$ 1.4; 65 were women, including 60 radiology and 63 non-radiology residents. Mean radiographic interpretation experience was 13.9 months among radiology residents and 14.7 months among non-radiology residents (mean difference, $-0.8$ months; 95\% CI, $-2.9$ to 1.2; $P = .43$). Of the 123 residents, 87 (70.7\%) were recruited from Center 1, 20 (16.3\%) from Center 2, and 16 (13.0\%) from Center 3 (Table~\ref{tab:characteristics}). Participant characteristics by support condition are shown in Supplementary Table~\ref{tab:s1}.

%Mean radiographic interpretation experience was similar between radiology and non-radiology residents (13.9 vs 14.7 months; mean difference, --0.8 months; 95\% CI, --2.9 to 1.2; $P = .43$). Of the 123 participants, 87 (70.7\%) were recruited from Center 1, 20 (16.3\%) from Center 2, and 16 (13.0\%) from Center 3 (Table~\ref{tab:characteristics}). Participant characteristics were similar across support conditions within each specialty stratum (Supplementary Table~\ref{tab:s1}).

\begin{table}[htbp]
\centering
\caption{Participant Characteristics Overall and by Specialty}
\label{tab:characteristics}
\begin{threeparttable}
\renewcommand{\arraystretch}{1.15}
\makebox[\textwidth][c]{\begin{tabular}{lccc}
\toprule
Characteristic & Total residents ($n = 123$) & Radiology ($n = 60$) & Non-radiology ($n = 63$) \\
\midrule
Age (y) & $24.1 \pm 1.4$ & $24.1 \pm 1.5$ & $24.0 \pm 1.3$ \\
Sex (male) & 58 (47.2) & 22 (36.7) & 36 (57.1) \\
Experience (m) & $14.3 \pm 5.7$ & $13.9 \pm 5.9$ & $14.7 \pm 5.5$ \\
Center & & & \\
\quad Center 1 & 87 (70.7) & 34 (56.7) & 53 (84.1) \\
\quad Center 2 & 20 (16.3) & 18 (30.0) & 2 (3.2) \\
\quad Center 3 & 16 (13.0) & 8 (13.3) & 8 (12.7) \\
\bottomrule
\end{tabular}}
\end{threeparttable}
\fulltablenote{Note.---Data are mean $\pm$ SD or $n$ (\%). Experience denotes months of clinical work involving radiographic interpretation. Center indicates the recruiting center.}
\end{table}

\subsection{Diagnostic Accuracy before and after AI Assistance}

\begin{figure}[H]
\centering
\includegraphics[width=\linewidth]{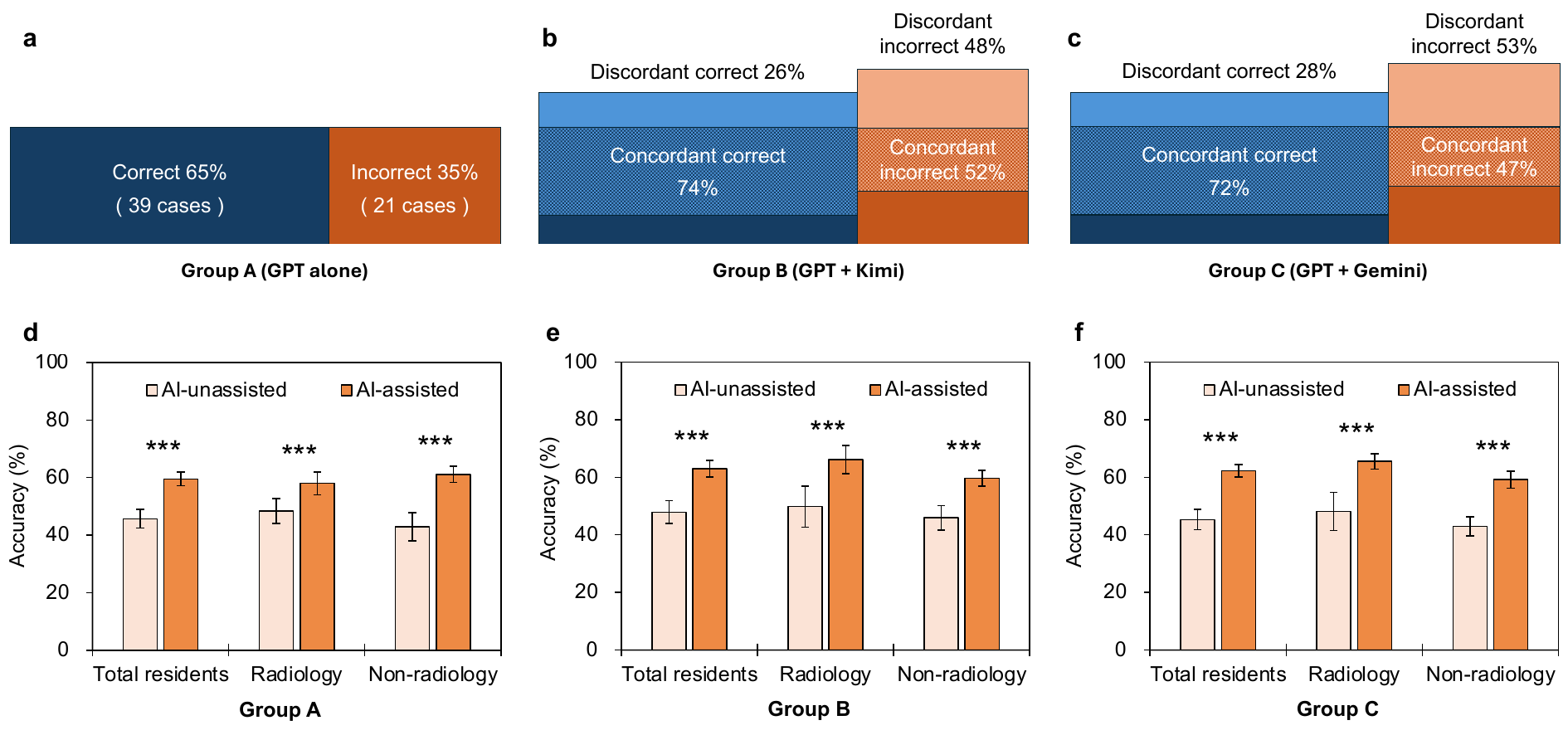}
\caption{\textbf{AI Model Concordance and Diagnostic Accuracy With and Without AI Assistance.} \textbf{a--c,} AI diagnostic performance and case-level concordance in groups A--C. In the dual-suggestion groups, concordant and discordant categories indicate whether the second model agreed with GPT-5.4 among GPT-5.4-correct and GPT-5.4-incorrect cases. \textbf{d--f,} Mean AI-unassisted and AI-assisted diagnostic accuracy among all residents, radiology residents, and non-radiology residents in groups A--C. Error bars represent 95\% CIs. ***$P < .001$ for paired comparisons between AI-unassisted and AI-assisted accuracy.}
\label{fig:3}
\end{figure}

The three AI models had identical overall diagnostic accuracy (65.0\%), while Kimi-K2.6 and Gemini-3.6 Flash differed in case-level concordance with GPT-5.4 (Figure~\ref{fig:3}a--c). Across all 123 residents, mean diagnostic accuracy increased from 46.3\% $\pm$ 11.3 to 61.5\% $\pm$ 8.0 after AI assistance (mean change, 15.2 percentage points; 95\% CI, 13.7--16.8; $P < .001$). Before AI assistance, radiology residents had higher accuracy than non-radiology residents (48.8\% vs 43.9\%; mean difference, 4.9 percentage points; 95\% CI, 0.9--8.9; $P = .017$). After AI assistance, accuracy remained higher among radiology residents (63.2\% vs 59.9\%; mean difference, 3.3 percentage points; 95\% CI, 0.4--6.1; $P = .024$), whereas accuracy change did not differ between specialties (14.4 vs 16.0 percentage points; mean difference, --1.6 percentage points; 95\% CI, --4.8 to 1.6; $P = .32$) (Supplementary Table~\ref{tab:s2}). Diagnostic accuracy increased after AI assistance in all three support conditions: by 13.9 percentage points in group A (95\% CI, 11.3--16.5), 15.0 percentage points in group B (95\% CI, 12.0--18.0), and 16.8 percentage points in group C (95\% CI, 14.1--19.6) (all $P < .001$) (Figure~\ref{fig:3}d--f; Supplementary Table~\ref{tab:s3}).

\subsection{Accuracy Change, Interpretation Time, and Confidence Change}

\begin{figure}[H]
\centering
\includegraphics[width=\linewidth]{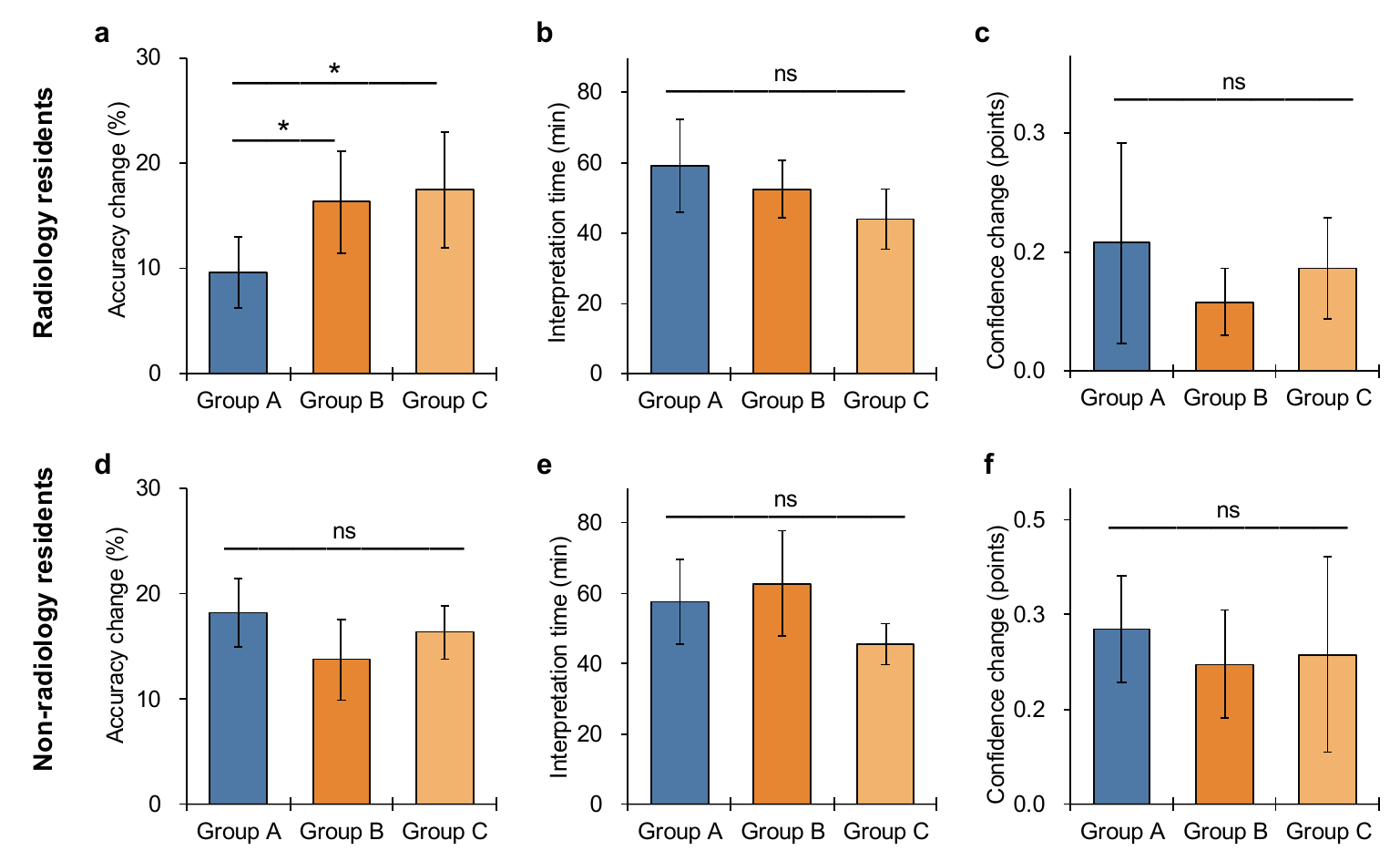}
\caption{\textbf{Diagnostic Outcomes by Support Condition.} \textbf{a--c,} Accuracy change, interpretation time, and confidence change, respectively, among radiology residents. \textbf{d--f,} Corresponding outcomes among non-radiology residents. Bars represent mean values, and error bars represent 95\% CIs. *Holm-adjusted $P < .05$ versus group A. ns indicates a nonsignificant overall comparison across the three support conditions.}
\label{fig:4}
\end{figure}

Among radiology residents, accuracy change differed across support conditions (group A, 9.6 percentage points $\pm$ 7.1; group B, 16.3 $\pm$ 10.6; group C, 17.5 $\pm$ 11.4; $P = .016$). Compared with group A, improvement was greater in group B by 6.69 percentage points (95\% CI, 0.97--12.40; Holm-adjusted $P = .030$) and in group C by 7.87 percentage points (95\% CI, 1.64--14.11; Holm-adjusted $P = .030$) (Figure~\ref{fig:4}a). Among non-radiology residents, accuracy change did not differ across support conditions (group A, 18.2 percentage points $\pm$ 6.9; group B, 13.7 $\pm$ 8.5; group C, 16.3 $\pm$ 5.6; $P = .20$) (Figure~\ref{fig:4}d). Interpretation time and confidence change did not differ across support conditions within either specialty stratum (all $P > .05$) (Figure~\ref{fig:4}; Supplementary Table~\ref{tab:s4}).

\subsection{Accuracy According to GPT-5.4 Suggestion Correctness}

\begin{figure}[H]
\centering
\includegraphics[width=\linewidth]{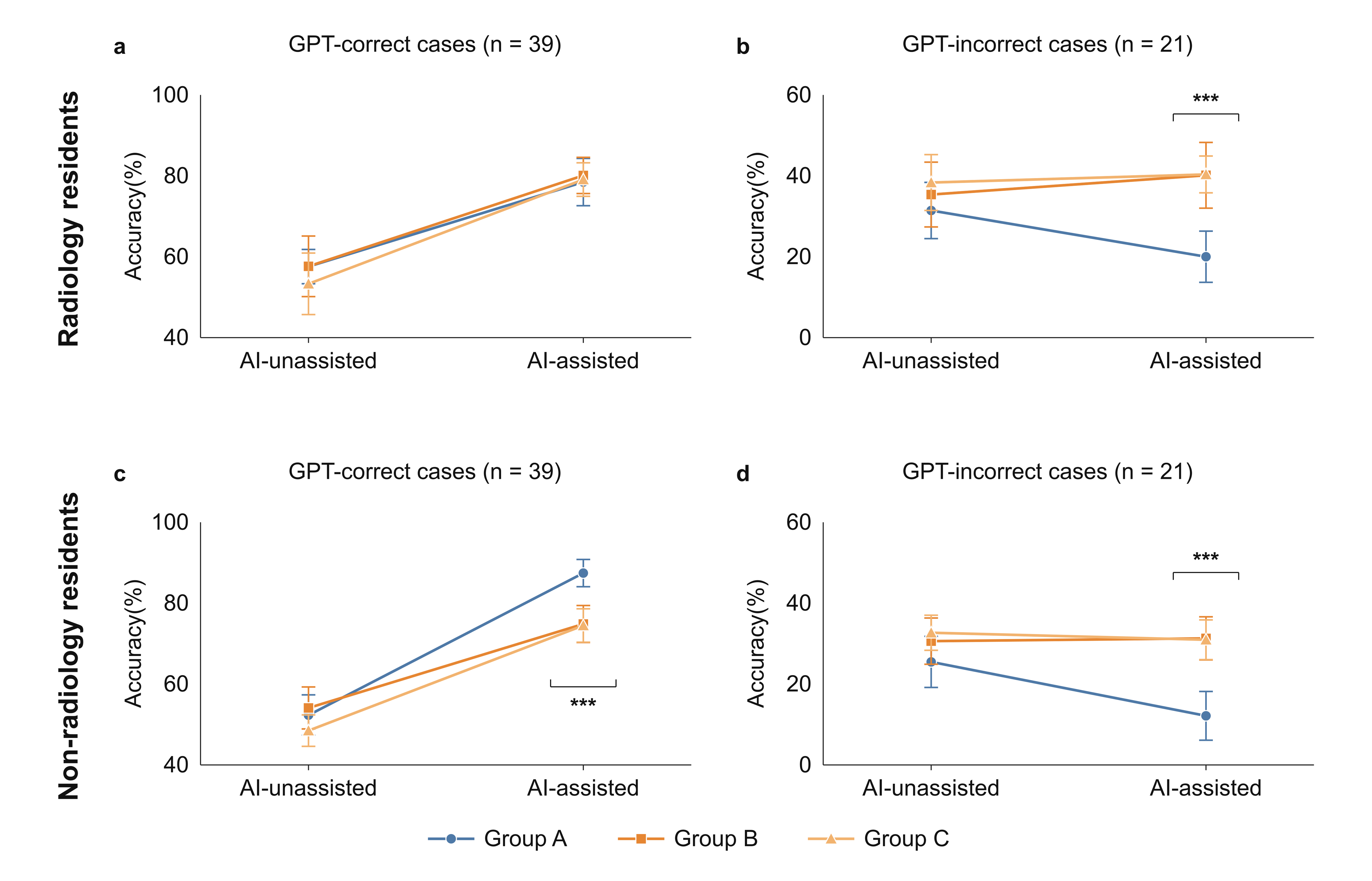}
\caption{\textbf{Diagnostic Accuracy According to GPT-5.4 Suggestion Correctness.} \textbf{a--b,} AI-unassisted and AI-assisted diagnostic accuracy among radiology residents in cases with correct ($n = 39$) and incorrect ($n = 21$) GPT-5.4 suggestions, respectively. \textbf{c--d,} Corresponding results among non-radiology residents. Lines represent groups A--C, and error bars represent 95\% CIs. ***Holm-adjusted $P < .001$ versus group A.}
\label{fig:5}
\end{figure}

Before AI assistance, accuracy did not differ across support conditions for GPT-5.4 correct or GPT-5.4 incorrect cases in either specialty (all $P > .05$). Among radiology residents, when GPT-5.4 was correct, AI-assisted accuracy did not differ across groups (78.5\% $\pm$ 12.5, 80.1\% $\pm$ 9.8, and 79.1\% $\pm$ 8.6, respectively; $P = .89$) (Figure~\ref{fig:5}a). When GPT-5.4 was incorrect, AI-assisted accuracy was higher in groups B and C than in group A (20.0\% $\pm$ 13.5, 40.1\% $\pm$ 17.9, and 40.4\% $\pm$ 9.4 for groups A, B, and C, respectively; both Holm-adjusted $P < .001$) (Figure~\ref{fig:5}b). Among non-radiology residents, when GPT-5.4 was correct, AI-assisted accuracy was lower in groups B and C than in group A (87.4\% $\pm$ 7.2, 74.8\% $\pm$ 10.0, and 74.5\% $\pm$ 9.3, respectively; both Holm-adjusted $P < .001$) (Figure~\ref{fig:5}c). In contrast, when GPT-5.4 was incorrect, AI-assisted accuracy was higher in groups B and C than in group A (12.1\% $\pm$ 12.9, 31.3\% $\pm$ 11.7, and 31.0\% $\pm$ 11.1, respectively; both Holm-adjusted $P < .001$) (Figure~\ref{fig:5}d) (Supplementary Tables~\ref{tab:s5}--\ref{tab:s7}).

\subsection{Revision Components and Specialty Interaction}

\begin{figure}[H]
\centering
\includegraphics[width=\linewidth]{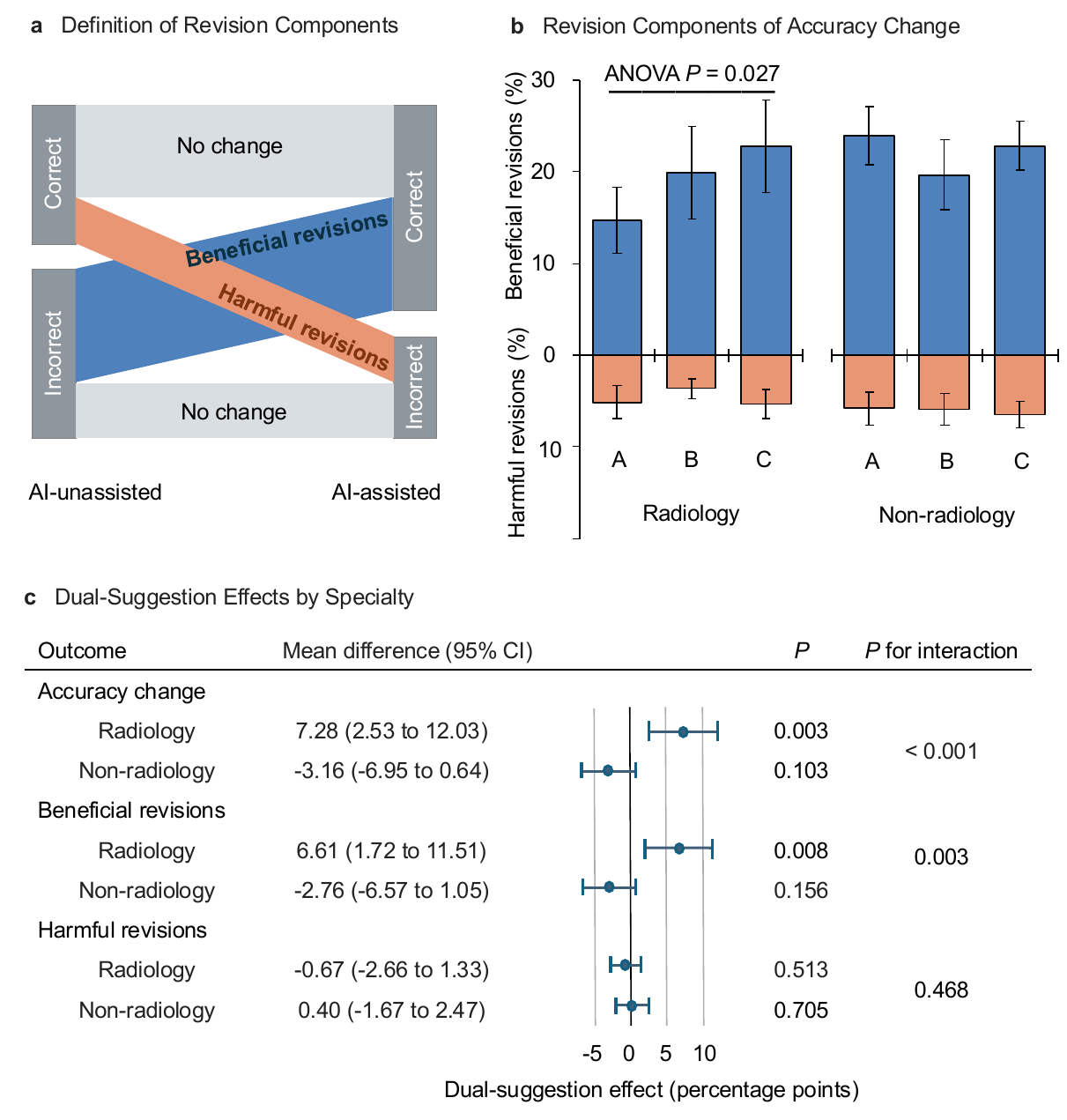}
\caption{\textbf{Revision Components and Dual-Suggestion Effects by Specialty.}
\textbf{a,} Definition of revision components. Beneficial and harmful revisions represent incorrect-to-correct and correct-to-incorrect transitions, respectively; gray flows indicate no change.
\textbf{b,} Revision components of accuracy change by support condition and specialty. Bars show means with 95\% CIs; harmful revisions are plotted below zero for visualization.
\textbf{c,} Dual-suggestion effects by specialty, defined as [(B+C)/2]--A. Points show mean differences with 95\% CIs; $P$ for interaction indicates differences between specialties.}
\label{fig:6}
\end{figure}

Accuracy change was decomposed into beneficial revisions (incorrect-to-correct) and harmful revisions (correct-to-incorrect). Among radiology residents, beneficial revisions differed across support conditions (group A, 14.8\% $\pm$ 7.8; group B, 19.9\% $\pm$ 11.0; group C, 22.8\% $\pm$ 10.5; $P = .027$), whereas harmful revisions did not (group A, 5.2\% $\pm$ 3.9; group B, 3.7\% $\pm$ 2.4; group C, 5.4\% $\pm$ 3.3; $P = .131$) (Figure~\ref{fig:6}a; Supplementary Table~\ref{tab:s8}). Among non-radiology residents, neither beneficial revisions ($P = .204$) nor harmful revisions ($P = .796$) differed across support conditions. The dual-suggestion effect on accuracy change differed by specialty
(interaction difference, 10.44 percentage points; 95\% CI, 4.36--16.52;
$P < .001$), with effects of 7.28 percentage points
(95\% CI, 2.53--12.03; $P = .003$) in radiology residents and
$-3.16$ percentage points (95\% CI, $-6.95$ to 0.64; $P = .103$)
in non-radiology residents. A specialty interaction was also observed
for beneficial revisions (9.37 percentage points; 95\% CI, 3.17--15.57;
$P = .003$), but not for harmful revisions ($-1.07$ percentage points;
95\% CI, $-3.95$ to 1.81; $P = .47$) (Figure~\ref{fig:6}b; Supplementary Table~\ref{tab:s9}). For GPT-5.4 incorrect cases, groups B and C showed higher beneficial revision rates than group A in both specialties (both $P < .001$; Supplementary Table~\ref{tab:s10}).

\section{Discussion}

In this prospective multicenter multireader study, we compared dual- with single-suggestion AI support for radiographic interpretation by radiology and non-radiology residents. When the shared GPT-5.4 suggestion was incorrect, AI-assisted accuracy was higher with dual-suggestion support in both groups. Radiology residents showed greater overall improvement in diagnostic accuracy with dual- than with single-suggestion support, whereas accuracy change did not differ across support conditions among non-radiology residents. Interpretation time did not differ across support conditions in either specialty. These findings suggest that dual-suggestion support may help limit the diagnostic influence of erroneous AI suggestions.

%This randomized multireader study evaluated whether dual-suggestion support could mitigate the diagnostic impact of erroneous AI suggestions among less experienced physicians during radiographic interpretation. Diagnostic accuracy improved after AI assistance across all three support conditions. When the shared GPT-5.4 suggestion was incorrect, AI-assisted accuracy was higher with dual- than single-suggestion support in both resident groups. Radiology residents showed greater improvement in diagnostic accuracy with dual- than single-suggestion support, whereas accuracy change did not differ across support conditions for non-radiology residents. Dual-suggestion support was not accompanied by longer interpretation time in either resident group. These findings suggest that dual-suggestion support may help limit the diagnostic impact of erroneous AI suggestions.

Previous studies have shown that AI assistance can improve diagnostic performance, while erroneous AI suggestions can also adversely influence diagnostic decisions\cite{jabbour2023mllm,chen2024impact}, particularly among less experienced physicians\cite{dratsch2023automation,teng2026overreliance}. Existing human–AI interaction strategies have largely focused on optimizing the presentation of information from a single AI. Our study extends this work by evaluating dual-suggestion support, in which an additional independent AI opinion is presented alongside the shared AI suggestion. The main contribution was observed when the shared AI suggestion was incorrect: diagnostic accuracy was higher with dual- than single-suggestion support in residents, suggesting that an additional independent AI opinion may help limit the diagnostic influence of erroneous AI suggestions. Importantly, this improvement was not accompanied by a detectable increase in interpretation time. These findings extend human–AI interaction research beyond single-AI presentation and suggest that dual-suggestion support may help limit the influence of erroneous AI suggestions without increasing interpretation time.

The higher diagnostic accuracy observed with dual-suggestion support when the shared AI suggestion was incorrect is consistent with evidence that access to an additional opinion can reduce inappropriate reliance on erroneous AI advice\cite{sokol2025artificial}. Outside medicine, actively seeking a second opinion has been shown to reduce overreliance on AI suggestions, including erroneous ones\cite{lu2024second}. In medicine, structured information sharing among physicians has been associated with fewer diagnostic errors\cite{centola2023networks}. These findings provide a conceptual basis for presenting an additional independent AI opinion when evaluating an AI-supported diagnosis. Notably, in an exploratory analysis of cases in which both AI suggestions were incorrect, dual- and single-suggestion support did not differ significantly in AI-assisted accuracy or increased confidence in incorrect final diagnoses (Supplementary Table~\ref{tab:s11}). Taken together, these findings support the potential value of dual-suggestion support when an AI suggestion may be erroneous.

The effect of dual-suggestion support differed by specialty. Although both groups had fewer than 3 years of clinical experience, radiology and non-radiology residents differed in specialty background and baseline diagnostic performance. Greater overall accuracy improvement with dual-suggestion support was observed in radiology residents but not in non-radiology residents. Among non-radiology residents, however, the effect varied with shared-AI correctness: accuracy was higher with dual-suggestion support when the shared AI suggestion was incorrect but lower when it was correct, while overall beneficial and harmful revision rates did not differ across support conditions. Previous studies have shown that the effect of AI assistance varies with task-specific expertise\cite{xu2026radiology,yu2024expertise}. Differences in domain-specific knowledge may influence how clinicians evaluate and accept AI suggestions\cite{schramm2026explanations,prinster2024explain,jackson2026domain}. In our study, specialty interactions were observed for accuracy change and beneficial revisions. The incremental effect of dual-suggestion support may vary across specialties.

Several limitations should be considered. First, process-level mechanisms were not directly assessed. Eye tracking can assess visual attention\cite{belde2026eyetracking,taib2026visual}, and suggestion-weighting measures can quantify the relative influence of dual AI suggestions on diagnostic decisions\cite{glickman2024weighting}. Future studies incorporating these measures may help clarify the mechanisms underlying the diagnostic value of dual-suggestion support. Second, the paired models differed in case-level error patterns, and dual-suggestion support therefore sometimes provided a correct additional suggestion when the shared GPT-5.4 suggestion was incorrect. The frequency and direction of such disagreement may vary across model pairings, and whether the effect of dual-suggestion support depends on these disagreement patterns warrants further investigation. Third, the 60-case set was purposively selected to balance disease composition and overall AI diagnostic accuracy and may not reflect the case mix of routine clinical practice. Further evaluation in unselected clinical populations is warranted. Nevertheless, the controlled multireader design enabled direct comparison of dual- and single-suggestion support under standardized conditions.

%Several limitations should be considered. First, the study included only less experienced physicians, limiting generalizability to more experienced clinicians. Second, process-level mechanisms remain to be characterized. Eye tracking can characterize visual attention and search behavior\cite{belde2026eyetracking,taib2026visual}, whereas direct assessment of suggestion weighting\cite{glickman2024weighting} can quantify the relative influence of dual AI suggestions on diagnostic decisions. Future studies should incorporate these measures to clarify the mechanisms underlying the diagnostic value of dual-suggestion support. Third, the two parallel dual-suggestion conditions showed consistency across two model pairings, but the range of model combinations remained limited. Whether the findings generalize to other model combinations or more than two simultaneous AI suggestions remains unknown. Fourth, the controlled case set was designed to isolate the effect of suggestion number rather than to estimate effectiveness under routine clinical case distributions. Validation in consecutive or prevalence-representative clinical cohorts is warranted.

In conclusion, dual-suggestion support may help limit the diagnostic influence of erroneous AI suggestions during radiographic interpretation. Compared with single-suggestion support, it was associated with greater overall improvement in diagnostic accuracy among radiology residents but not among non-radiology residents, without increased interpretation time. These findings highlight the potential of human–AI interaction design to reduce the diagnostic consequences of erroneous AI suggestions.

%In conclusion, dual-suggestion support may mitigate the diagnostic influence of erroneous AI suggestions in less experienced physicians. Compared with single-suggestion support, dual-suggestion support showed greater overall improvement in diagnostic accuracy among radiology residents but not among non-radiology residents, with no detectable differences in interpretation time across support conditions. Dual-suggestion support may therefore be considered as a human–AI interaction strategy for radiographic interpretation, particularly among less experienced radiology residents.

\section*{Data Availability}
% The study was approved by the Ethics Committee of the First Affiliated Hospital of Nanchang University (2026-756) and was registered with the Chinese Clinical Trial Registry (ChiCTR2600129243). 
The deidentified 60-case radiographs, groud-truth diagnoses, and the diagnostic lexicon used for free-text scoring are available at \href{https://github.com/zhexu1997/prospective-multicenter-ai-support-for-radiographic-interpretation}{GitHub}.

\section*{Code Availability}
The inference code is available at \href{https://github.com/zhexu1997/prospective-multicenter-ai-support-for-radiographic-interpretation}{GitHub}. Model outputs were generated through platform APIs: GPT-5.4 and Kimi-K2.6 via \href{https://ai.azure.com}{Microsoft Foundry} (formerly Azure AI Foundry), and Gemini-3.6 Flash via \href{https://cloud.google.com/vertex-ai}{Google Vertex AI}.

\section*{Funding}
This study was supported by the Jiangxi Province Natural Science Foundation (No. 20232BAB216099), the Jiangxi Clinical Research Center for Medical Imaging (No. 20223BCG74001), and the Jiangxi Province Key Laboratory for Precision Pathology and Intelligent Diagnosis (No. 2024SSY06281).

%\section*{Author Contributions}
%Guarantors of integrity of entire study, L.W., H.C., F.Q.Z.; study concepts/study design, L.W., Z.X., H.Y.W., H.C., F.Q.Z.; data acquisition, L.W., F.F.Z., W.D., Y.Y.C.; case-difficulty assessment, C.L.Z., Y.T.Z., K.X.C., X.L., C.Y.; response adjudication and validation, L.W., W.D.; software and technical implementation, Z.X., H.Y.W., H.C.; data analysis/interpretation, L.W., Z.X., H.Y.W., H.C., F.Q.Z.; statistical analysis, L.W., Z.X., H.Y.W.; manuscript drafting, L.W.; manuscript revision for important intellectual content, all authors; approval of final version of submitted manuscript, all authors.  

% RSNA / NLM reference list. Do not run BibTeX on this file; naturemag-doi.bst
% does not match Radiology: Artificial Intelligence style.
% Radiology: Artificial Intelligence reference style (NLM / AMA).
% https://pubs.rsna.org/page/ai/author-instructions
% Six or fewer authors are listed in full. Seven or more: first three, then et al.
% Journal abbreviations follow the NLM catalog. DOIs are included.

\clearpage
% =====================================================================
% SUPPLEMENTARY MATERIALS (same file)
% =====================================================================
\setcounter{figure}{0}
\setcounter{table}{0}
\renewcommand{\thefigure}{S\arabic{figure}}
\renewcommand{\thetable}{S\arabic{table}}
\raggedbottom

\section*{Supplementary Materials}

These tables provide participant-level descriptive results, prespecified support-condition comparisons, and additional analyses for the 123 participants included in the study.

Group A: GPT; group B: GPT plus Kimi; group C: GPT plus Gemini. Radiology residents: $n = 60$ (A/B/C, 20/21/19); non-radiology residents: $n = 63$ (20/21/22).

All groups evaluated the same 60 cases. Conditional analyses used the 39 cases in which the common GPT suggestion was correct and the 21 cases in which it was incorrect. This classification was identical across groups.

Accuracy change is AI-assisted minus AI-unassisted accuracy. Beneficial and harmful revisions are incorrect-to-correct and correct-to-incorrect transitions, respectively. Transition percentages use all 60 cases as the denominator. Interpretation time sums both interpretation stages; confidence change is final minus initial mean confidence.

\subsection*{Contents}
\begin{itemize}[leftmargin=1.4em,itemsep=2pt]
\item Supplementary Statistical Methods
\item Supplementary Appendix 1. Case Composition, Model Concordance, and Representative Fixed AI Suggestions
\item Figure S1. Composition of the 60-case set
\item Figure S2. Correctness-status maps for the study conditions
\item Supplementary Appendix 2. Prespecified Diagnostic Lexicon for Free-Text Response Scoring
\item Appendix Table A1. Accepted Diagnostic Expressions and Examples Not Accepted as Standalone Answers
\item Supplementary Table S1. Participant Characteristics by Specialty and Support Condition
\item Supplementary Table S2. Participant-Level Outcomes by Specialty
\item Supplementary Table S3. Diagnostic Accuracy Before and After AI Assistance by Specialty and Support Condition
\item Supplementary Table S4. Accuracy Change, Interpretation Time, and Confidence Change by Specialty and Support Condition
\item Supplementary Table S5. Diagnostic Accuracy in Cases With a Correct Common GPT Suggestion
\item Supplementary Table S6. Diagnostic Accuracy in Cases With an Incorrect Common GPT Suggestion
\item Supplementary Table S7. Pairwise Accuracy Contrasts by Specialty and GPT Suggestion Correctness
\item Supplementary Table S8. Diagnostic Revision Patterns by Specialty and Support Condition
\item Supplementary Table S9. Dual-Suggestion Effects and Specialty Interactions
\item Supplementary Table S10. Revision Components by Shared GPT Suggestion Correctness, Specialty, and Support Condition
\item Supplementary Table S11. Outcomes in Cases for Which Both AI Models Were Incorrect
\end{itemize}

\clearpage
\subsection*{Supplementary Methods: Allocation Procedure}

Participant allocation was performed within specialty strata by an investigator not involved in participant assessment. Deidentified participant information, including specialty classification, was submitted to DeepSeek using a prespecified prompt instructing random and balanced allocation in a 1:1:1 ratio to groups A, B, and C. The resulting allocation list was accepted without manual modification, fixed before participant assessment, and concealed from enrollment personnel until assignment. DeepSeek was used solely for participant allocation and was not one of the diagnostic AI models evaluated in the study.

\subsection*{Supplementary Statistical Methods}

All analyses used participant-level outcomes. Paired $t$ tests assessed AI-unassisted versus AI-assisted accuracy within each group. Welch $t$ tests compared specialties, and Welch ANOVA assessed differences across support conditions within each specialty. Prespecified B--A and C--A comparisons used Welch $t$ tests with Holm adjustment within each outcome and specialty stratum; for conditional analyses, adjustment was additionally performed within each GPT-correctness subset. B--C comparisons were exploratory and unadjusted.

Additional analyses compared AI-unassisted accuracy, AI-assisted accuracy, and accuracy change across support conditions separately within the shared GPT-correct and GPT-incorrect subsets (Supplementary Tables~\ref{tab:s5}--\ref{tab:s7}). Separate ordinary least-squares models with heteroskedasticity-robust standard errors using the HC3 estimator evaluated accuracy change, beneficial revisions, and harmful revisions. Dual-suggestion effect was the equally weighted contrast $[(B + C)/2] - A$. B--A and C--A effects were also estimated separately within each specialty. Specialty interaction contrasts were the radiology effect minus the corresponding non-radiology effect. Individual contrasts used two-sided Wald $z$ tests. The overall three-group-by-specialty interaction used a two-degree-of-freedom Wald chi-square test. Model-specific B--A and C--A contrasts were Holm-adjusted in pairs within each outcome and analysis block. 
An additional exploratory analysis evaluated cases in which both AI suggestions were incorrect (Supplementary Table~\ref{tab:s11}). For the GPT plus Kimi condition, group B was compared with group A using the 11 cases in which both GPT-5.4 and Kimi-K2.6 were incorrect; for the GPT plus Gemini condition, group C was compared with group A using the 10 cases in which both GPT-5.4 and Gemini-3.6 Flash were incorrect. Participant-level outcomes included AI-assisted accuracy, accuracy change, confidence change, and the proportion of incorrect final diagnoses with increased confidence. Comparisons used two-sided Welch $t$ tests, with Holm adjustment applied to the two dual-versus-single comparisons within each outcome.
All tests were two-sided, with $P < .05$ indicating statistical significance. Analyses were performed using Python, version 3.12.14.

\clearpage
\subsection*{Supplementary Appendix 1. Case Composition, Model Concordance, and Representative Fixed AI Suggestions}

This appendix describes the 60-case set, the correctness patterns of the AI suggestions, and four representative fixed outputs. Correctness refers to the primary diagnosis under the diagnostic lexicon.

\paragraph{Disease composition.}
Thirty examinations were chest radiographs and 30 were abdominal radiographs. Urinary tract calculi include renal and ureteral calculi. Normal abdominal examinations are shown separately from disease categories.

\begin{figure}[H]
\centering
\includegraphics[width=\linewidth]{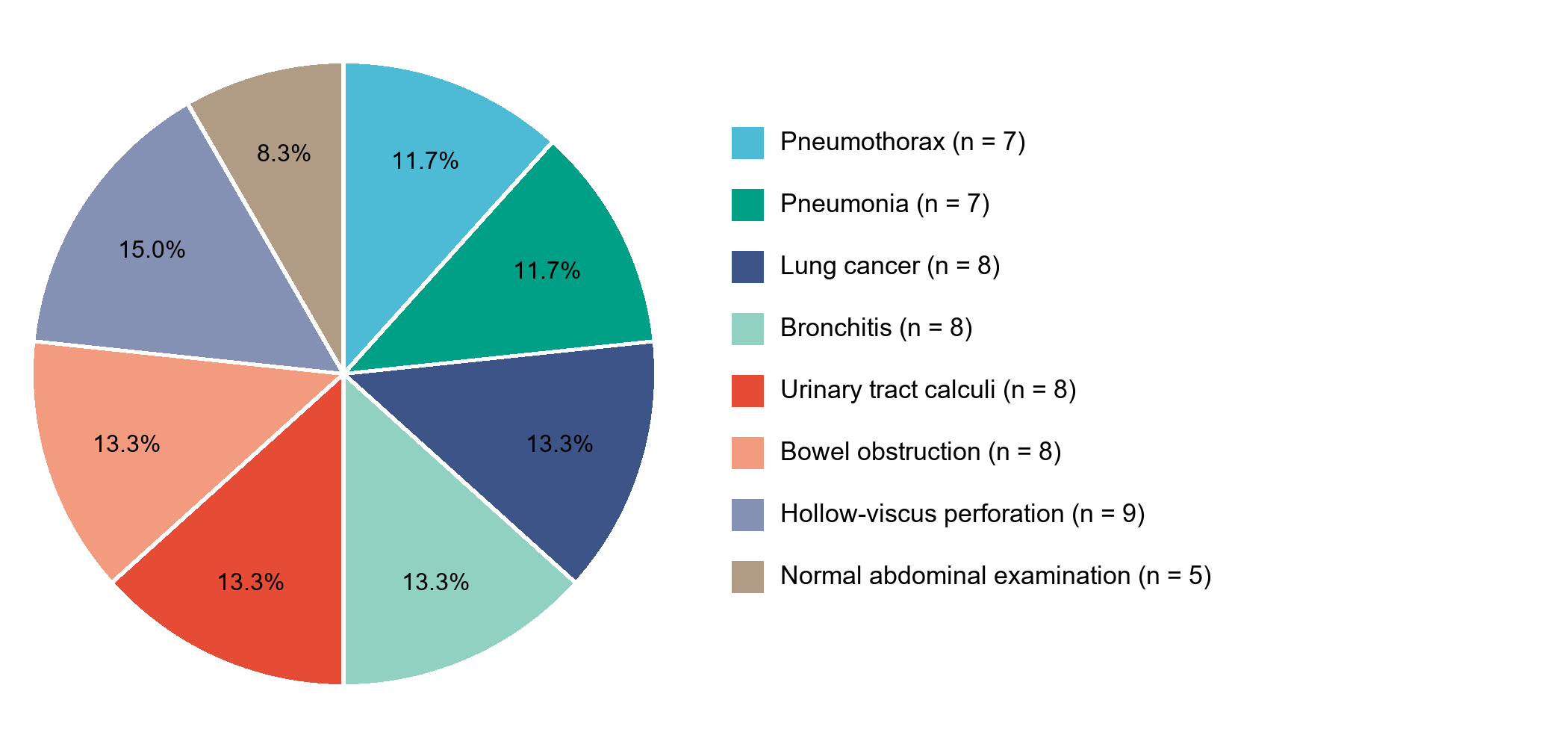}
\caption{\textbf{Composition of the 60-case set.} Thirty examinations were chest radiographs and 30 were abdominal radiographs. Urinary tract calculi include renal and ureteral calculi. Normal abdominal examinations are shown separately from disease categories.}
\label{fig:s1}
\end{figure}

\paragraph{Case selection and control of model performance.} Model outputs for all 218 candidate examinations were generated and fixed before selection of the final study cases. The final 60-case set was selected to balance disease composition and to match overall diagnostic accuracy across GPT-5.4, Kimi-K2.6, and Gemini-3.6 Flash. Thirty chest and 30 abdominal examinations were included, and each model correctly diagnosed 39 of 60 cases (65.0\%). Matching overall accuracy did not require the models to be correct on the same cases, thereby preserving case-level differences in model correctness and concordance.

%Because the study was designed to isolate the effect of suggestion number rather than differences in underlying model performance, the final case set was selected to equalize overall diagnostic accuracy across the three AI models. Each model correctly diagnosed 39 of 60 cases (65.0\%). Matching overall accuracy did not require the models to be correct on the same cases.

\paragraph{Case-level correctness concordance.}
Each square represents one case, numbered 1--60 in the same order in both panels. Example case numbers correspond to these square labels. Concordance means agreement in correctness status, not agreement in diagnostic wording.

\textbf{Group A (GPT).} GPT correct: 39; GPT incorrect: 21. Group A has no second-model status.

\textbf{Group B (GPT plus Kimi).} Both correct: 29; GPT correct / second model incorrect: 10; GPT incorrect / second model correct: 10; both incorrect: 11.

\textbf{Group C (GPT plus Gemini).} Both correct: 28; GPT correct / second model incorrect: 11; GPT incorrect / second model correct: 11; both incorrect: 10.

\begin{figure}[H]
\centering
\begingroup
\setlength{\tabcolsep}{3pt}
\renewcommand{\arraystretch}{1.35}
\textbf{Group A (GPT)}\\[4pt]
\begin{tabular}{ll}
\colorbox{sugcorrect}{\makebox[9.1em]{\strut\textcolor{white}{}}} & GPT correct ($n = 39$) \\
\colorbox{sugwrong}{\makebox[4.9em]{\strut\textcolor{white}{}}} & GPT incorrect ($n = 21$) \\
\end{tabular}\\[20pt]
\textbf{Group B (GPT plus Kimi)}\\[4pt]
\begin{tabular}{*{10}{c}}
\cellcolor{bothincorrect}1 & \cellcolor{bothincorrect}2 & \cellcolor{bothincorrect}3 & \cellcolor{bothincorrect}4 & \cellcolor{bothincorrect}5 & \cellcolor{bothincorrect}6 & \cellcolor{gptincseccorr}\color{white}7 & \cellcolor{bothcorrect}\color{white}8 & \cellcolor{bothcorrect}\color{white}9 & \cellcolor{bothcorrect}\color{white}10 \\
\cellcolor{bothcorrect}\color{white}11 & \cellcolor{gptincseccorr}\color{white}12 & \cellcolor{gptcorrsecinc}13 & \cellcolor{bothcorrect}\color{white}14 & \cellcolor{bothcorrect}\color{white}15 & \cellcolor{bothcorrect}\color{white}16 & \cellcolor{bothincorrect}17 & \cellcolor{bothcorrect}\color{white}18 & \cellcolor{bothcorrect}\color{white}19 & \cellcolor{bothcorrect}\color{white}20 \\
\cellcolor{bothcorrect}\color{white}21 & \cellcolor{bothcorrect}\color{white}22 & \cellcolor{bothcorrect}\color{white}23 & \cellcolor{bothcorrect}\color{white}24 & \cellcolor{bothcorrect}\color{white}25 & \cellcolor{gptcorrsecinc}26 & \cellcolor{gptcorrsecinc}27 & \cellcolor{bothcorrect}\color{white}28 & \cellcolor{gptincseccorr}\color{white}29 & \cellcolor{bothincorrect}30 \\
\cellcolor{bothcorrect}\color{white}31 & \cellcolor{gptcorrsecinc}32 & \cellcolor{bothcorrect}\color{white}33 & \cellcolor{bothcorrect}\color{white}34 & \cellcolor{gptcorrsecinc}35 & \cellcolor{gptincseccorr}\color{white}36 & \cellcolor{gptcorrsecinc}37 & \cellcolor{bothcorrect}\color{white}38 & \cellcolor{bothincorrect}39 & \cellcolor{bothcorrect}\color{white}40 \\
\cellcolor{bothincorrect}41 & \cellcolor{bothcorrect}\color{white}42 & \cellcolor{bothcorrect}\color{white}43 & \cellcolor{bothcorrect}\color{white}44 & \cellcolor{gptcorrsecinc}45 & \cellcolor{bothcorrect}\color{white}46 & \cellcolor{gptincseccorr}\color{white}47 & \cellcolor{gptincseccorr}\color{white}48 & \cellcolor{bothcorrect}\color{white}49 & \cellcolor{bothcorrect}\color{white}50 \\
\cellcolor{gptincseccorr}\color{white}51 & \cellcolor{gptincseccorr}\color{white}52 & \cellcolor{bothcorrect}\color{white}53 & \cellcolor{gptincseccorr}\color{white}54 & \cellcolor{gptincseccorr}\color{white}55 & \cellcolor{gptcorrsecinc}56 & \cellcolor{bothincorrect}57 & \cellcolor{bothcorrect}\color{white}58 & \cellcolor{gptcorrsecinc}59 & \cellcolor{gptcorrsecinc}60 \\
\end{tabular}\\[20pt]
\textbf{Group C (GPT plus Gemini)}\\[4pt]
\begin{tabular}{*{10}{c}}
\cellcolor{bothincorrect}1 & \cellcolor{bothincorrect}2 & \cellcolor{bothincorrect}3 & \cellcolor{bothincorrect}4 & \cellcolor{bothincorrect}5 & \cellcolor{bothincorrect}6 & \cellcolor{gptincseccorr}\color{white}7 & \cellcolor{bothcorrect}\color{white}8 & \cellcolor{bothcorrect}\color{white}9 & \cellcolor{bothcorrect}\color{white}10 \\
\cellcolor{bothcorrect}\color{white}11 & \cellcolor{gptincseccorr}\color{white}12 & \cellcolor{gptcorrsecinc}13 & \cellcolor{bothcorrect}\color{white}14 & \cellcolor{bothcorrect}\color{white}15 & \cellcolor{bothcorrect}\color{white}16 & \cellcolor{bothincorrect}17 & \cellcolor{gptcorrsecinc}18 & \cellcolor{gptcorrsecinc}19 & \cellcolor{gptcorrsecinc}20 \\
\cellcolor{gptcorrsecinc}21 & \cellcolor{gptcorrsecinc}22 & \cellcolor{gptcorrsecinc}23 & \cellcolor{gptcorrsecinc}24 & \cellcolor{bothcorrect}\color{white}25 & \cellcolor{bothcorrect}\color{white}26 & \cellcolor{bothcorrect}\color{white}27 & \cellcolor{bothcorrect}\color{white}28 & \cellcolor{gptincseccorr}\color{white}29 & \cellcolor{gptincseccorr}\color{white}30 \\
\cellcolor{bothcorrect}\color{white}31 & \cellcolor{gptcorrsecinc}32 & \cellcolor{bothcorrect}\color{white}33 & \cellcolor{bothcorrect}\color{white}34 & \cellcolor{gptcorrsecinc}35 & \cellcolor{bothincorrect}36 & \cellcolor{gptcorrsecinc}37 & \cellcolor{bothcorrect}\color{white}38 & \cellcolor{bothincorrect}39 & \cellcolor{bothcorrect}\color{white}40 \\
\cellcolor{bothincorrect}41 & \cellcolor{bothcorrect}\color{white}42 & \cellcolor{bothcorrect}\color{white}43 & \cellcolor{bothcorrect}\color{white}44 & \cellcolor{bothcorrect}\color{white}45 & \cellcolor{bothcorrect}\color{white}46 & \cellcolor{gptincseccorr}\color{white}47 & \cellcolor{gptincseccorr}\color{white}48 & \cellcolor{bothcorrect}\color{white}49 & \cellcolor{bothcorrect}\color{white}50 \\
\cellcolor{gptincseccorr}\color{white}51 & \cellcolor{gptincseccorr}\color{white}52 & \cellcolor{bothcorrect}\color{white}53 & \cellcolor{gptincseccorr}\color{white}54 & \cellcolor{gptincseccorr}\color{white}55 & \cellcolor{bothcorrect}\color{white}56 & \cellcolor{gptincseccorr}\color{white}57 & \cellcolor{bothcorrect}\color{white}58 & \cellcolor{bothcorrect}\color{white}59 & \cellcolor{bothcorrect}\color{white}60 \\
\end{tabular}
\endgroup
\vspace{20pt}

\begin{tabular}{ll}
\colorbox{bothcorrect}{\makebox[1.1em]{\strut\textcolor{white}{}}} & Both correct \\
\colorbox{gptcorrsecinc}{\makebox[1.1em]{\strut}} & GPT correct / second model incorrect \\
\colorbox{gptincseccorr}{\makebox[1.1em]{\strut\textcolor{white}{}}} & GPT incorrect / second model correct \\
\colorbox{bothincorrect}{\makebox[1.1em]{\strut}} & Both incorrect \\
\end{tabular}

\caption{\textbf{Correctness-status maps for the single- and dual-suggestion conditions.} Each square represents one case, numbered 1--60 in the same order in both dual-suggestion panels. Group A has no second-model status and uses the same two-color key as the examples below (teal, correct; rust, incorrect). The four colors below apply only to groups B and C. Both-incorrect cases count as correctness-concordant even when the diagnostic labels differ.}
\label{fig:s2}
\end{figure}

\clearpage
The four examples below use the same case-level correctness categories as Figure~\ref{fig:s2}. Teal panels mark a correct primary diagnosis; rust panels mark an incorrect primary diagnosis.

\begin{tcolorbox}[casesbox, title={Example 1. Case 8 --- GPT correct / Kimi correct}]
\casemeta{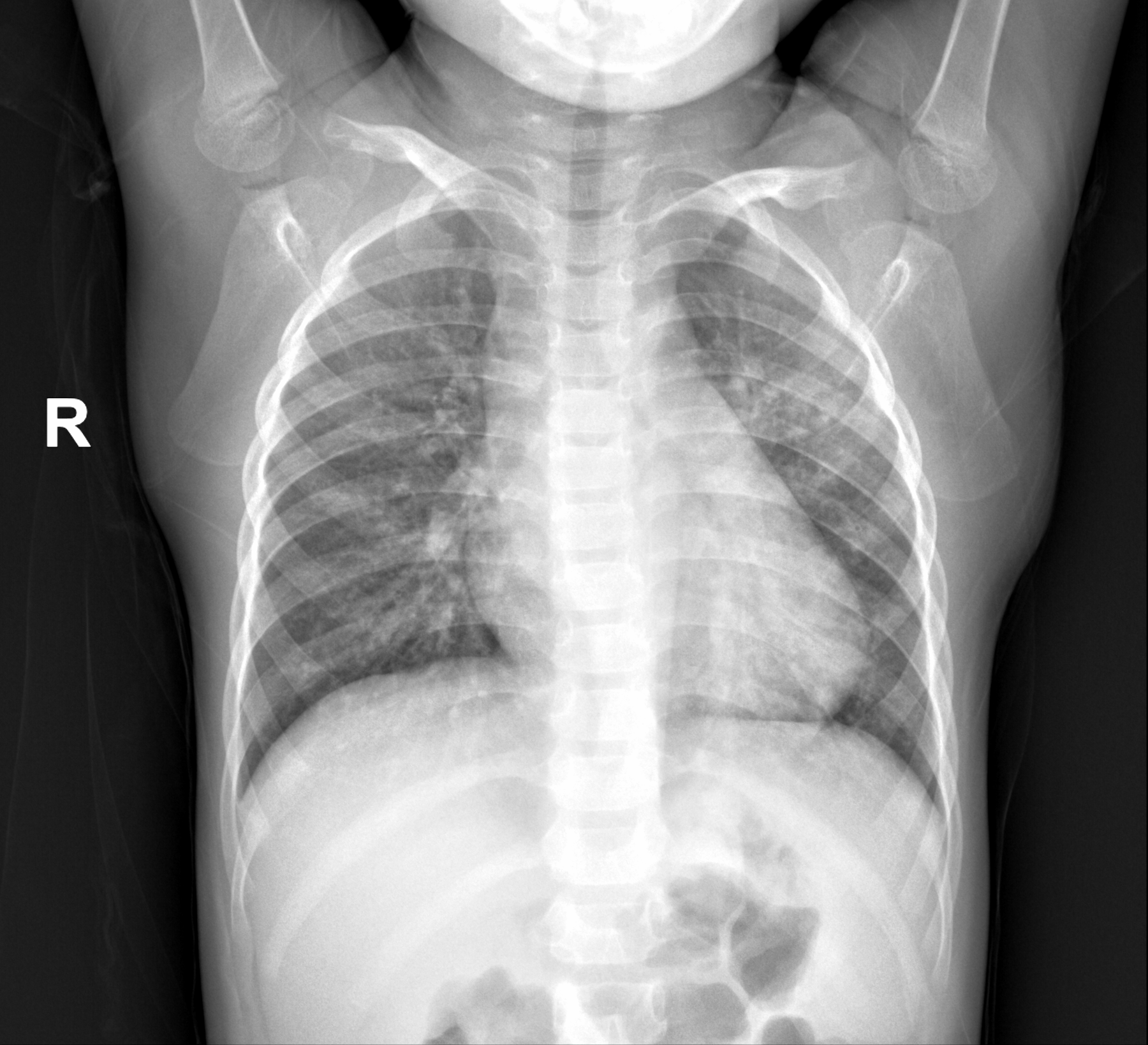}{G0385195}{Pneumonia}{Cough and chest tightness.}{both models correct}

\vspace{8pt}
\begin{tcbraster}[raster columns=2, raster equal height=rows, raster column skip=1em]
\begin{tcolorbox}[modelbox, title={GPT suggestion (correct)}, colback=sugcorrect!8, colframe=sugcorrect, colbacktitle=sugcorrect]
\textbf{Primary diagnosis:} Bronchopneumonia\\[3pt]
The chest radiograph shows patchy and reticular ill-defined opacities around both hila and in both lower lungs, with increased lung markings, more pronounced in the left middle and lower lung. The heart is not enlarged, and there is no pleural effusion. With cough and chest tightness, bronchopneumonia is the first consideration, followed by acute bronchitis.
\end{tcolorbox}
\begin{tcolorbox}[modelbox, title={Kimi suggestion (correct)}, colback=sugcorrect!8, colframe=sugcorrect, colbacktitle=sugcorrect]
\textbf{Primary diagnosis:} Bronchopneumonia\\[3pt]
The chest radiograph shows increased and ill-defined lung markings bilaterally, scattered patchy ill-defined opacities, and increased hilar opacity. With cough and chest tightness, the findings are consistent with bronchopneumonia and airway inflammatory changes.
\end{tcolorbox}
\end{tcbraster}
\end{tcolorbox}

\clearpage

\begin{tcolorbox}[casesbox, title={Example 2. Case 13 --- GPT correct / Kimi incorrect}]
\casemeta{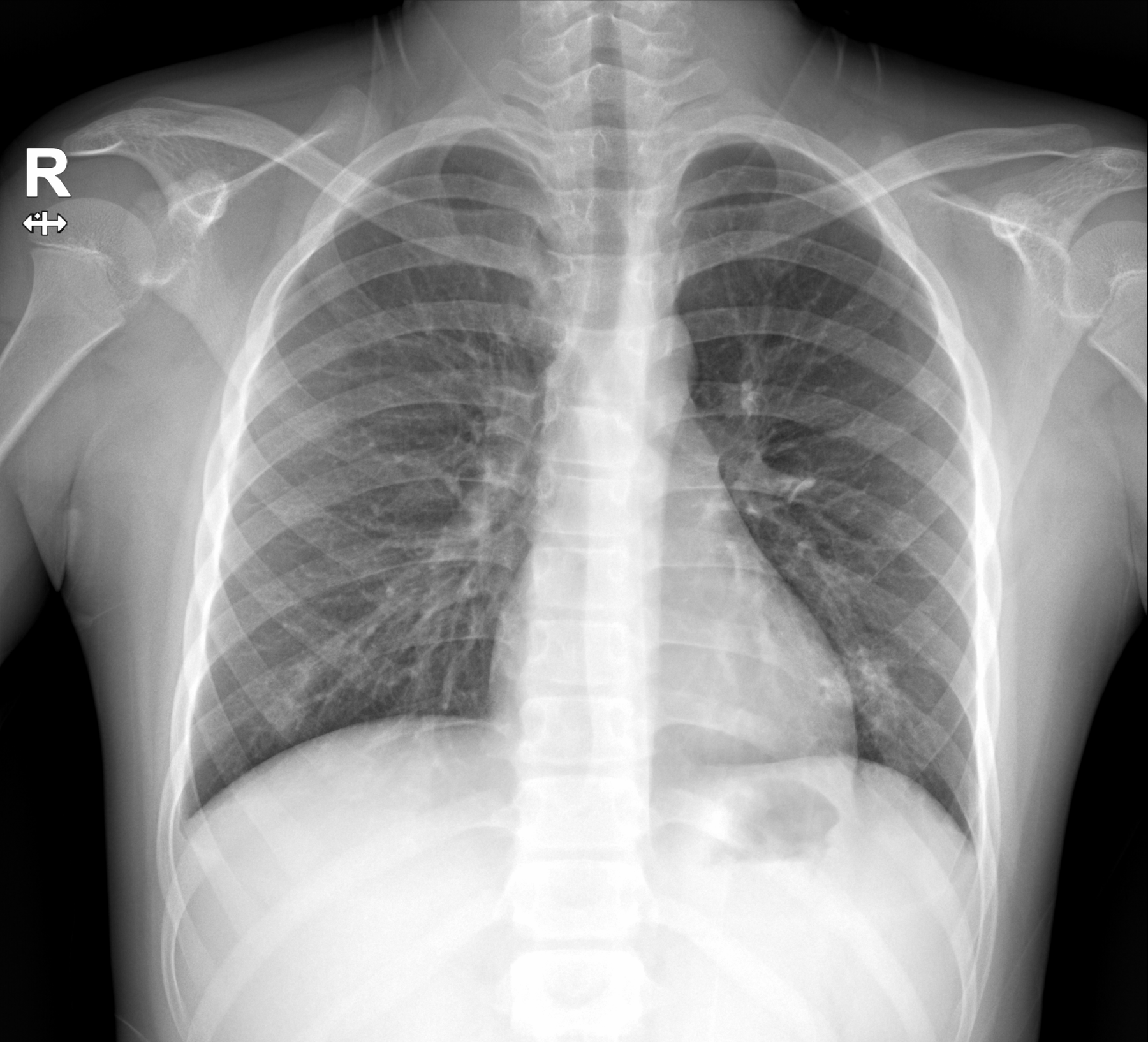}{G0404250}{Pneumonia}{Cough and chest tightness.}{GPT correct / second model incorrect}

\vspace{8pt}
\begin{tcbraster}[raster columns=2, raster equal height=rows, raster column skip=1em]
\begin{tcolorbox}[modelbox, title={GPT suggestion (correct)}, colback=sugcorrect!8, colframe=sugcorrect, colbacktitle=sugcorrect]
\textbf{Primary diagnosis:} Left lower-lobe pneumonia\\[3pt]
The chest radiograph shows patchy consolidation in the left lower lung and retrocardiac region, with mildly increased lung markings bilaterally. The heart is not enlarged, and there is no pleural effusion. With cough and chest tightness, left lower-lung infection with associated bronchitic changes is the first consideration.
\end{tcolorbox}
\begin{tcolorbox}[modelbox, title={Kimi suggestion (incorrect)}, colback=sugwrong!8, colframe=sugwrong, colbacktitle=sugwrong]
\textbf{Primary diagnosis:} Acute bronchitis\\[3pt]
The chest radiograph shows increased and thickened lung markings with ill-defined margins, without evident consolidation or pleural effusion. With cough and chest tightness, the imaging appearance is consistent with bronchitis. The cardiac silhouette and diaphragm are unremarkable.
\end{tcolorbox}
\end{tcbraster}
\end{tcolorbox}

\clearpage

\begin{tcolorbox}[casesbox, title={Example 3. Case 7 --- GPT incorrect / Kimi correct}]
\casemeta{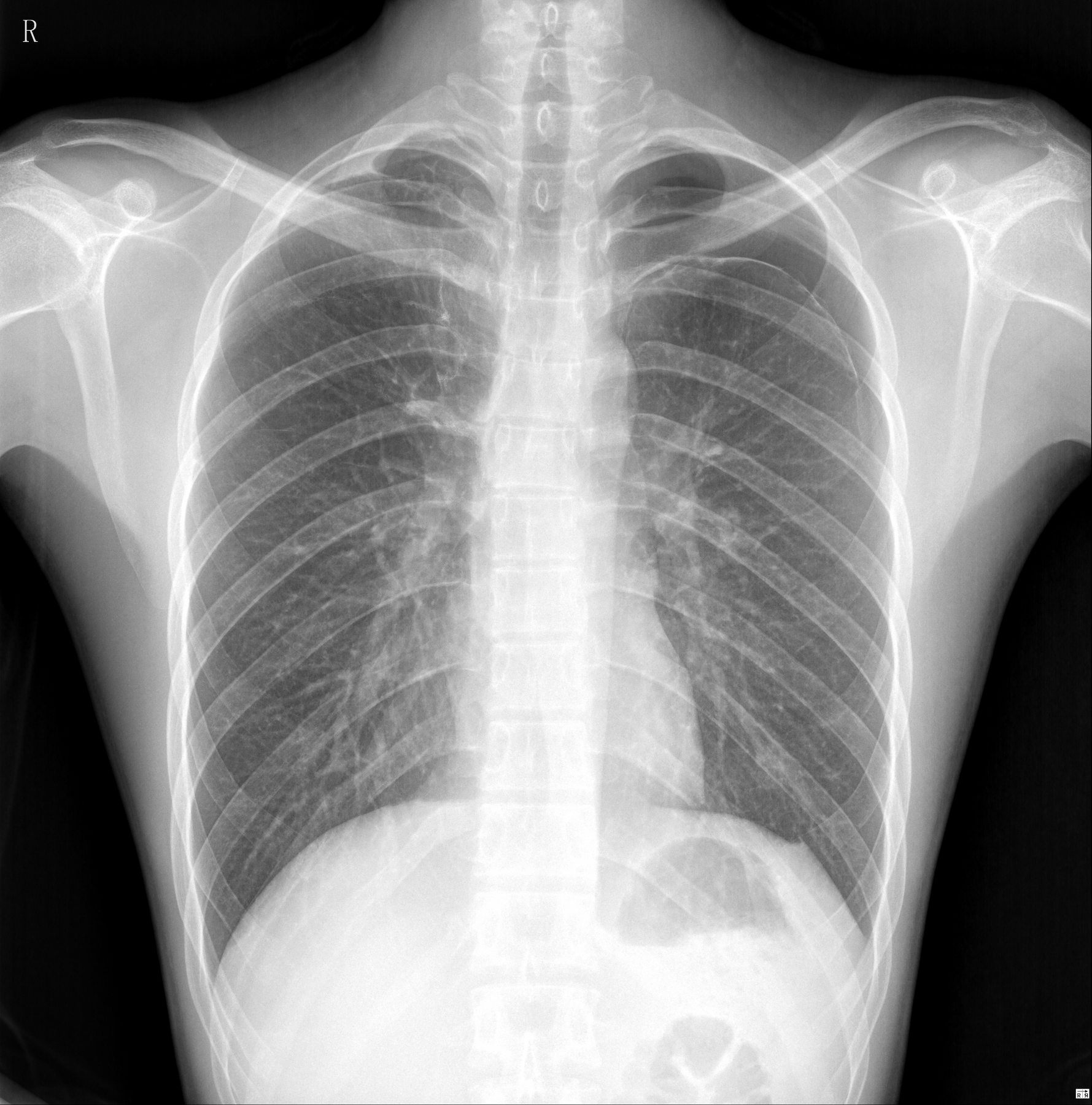}{P4216711}{Pneumothorax}{Cough and chest tightness.}{GPT incorrect / second model correct}

\vspace{8pt}
\begin{tcbraster}[raster columns=2, raster equal height=rows, raster column skip=1em]
\begin{tcolorbox}[modelbox, title={GPT suggestion (incorrect)}, colback=sugwrong!8, colframe=sugwrong, colbacktitle=sugwrong]
\textbf{Primary diagnosis:} Bronchitis / small-airway inflammation\\[3pt]
The chest radiograph shows mildly increased bilateral lung lucency and mildly increased and thickened markings in the hilar and lower-lung regions. There is no focal consolidation, pleural effusion, or pneumothorax. The heart is not enlarged. With cough and chest tightness, bronchitis or small-airway inflammation is favored.
\end{tcolorbox}
\begin{tcolorbox}[modelbox, title={Kimi suggestion (correct)}, colback=sugcorrect!8, colframe=sugcorrect, colbacktitle=sugcorrect]
\textbf{Primary diagnosis:} Left pneumothorax\\[3pt]
The chest radiograph shows abnormal lucency without lung markings in the peripheral left upper lung, with a visible visceral pleural line and the margin of the compressed lung. With cough and chest tightness, the findings are consistent with left pneumothorax.
\end{tcolorbox}
\end{tcbraster}
\end{tcolorbox}

\clearpage

\begin{tcolorbox}[casesbox, title={Example 4. Case 1 --- GPT incorrect / Kimi incorrect}]
\casemeta{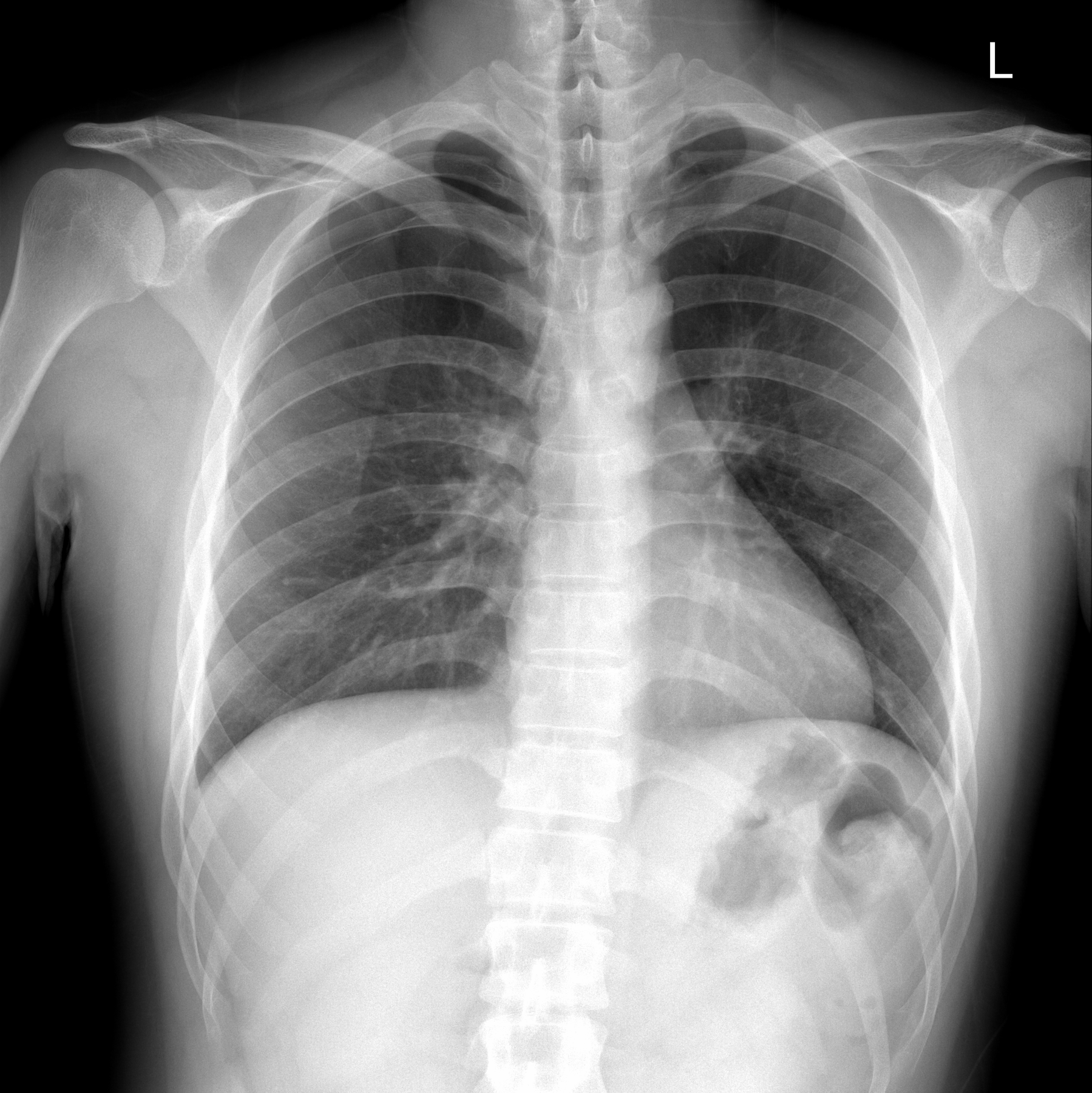}{P537136}{Pneumothorax}{Cough and chest tightness.}{both models incorrect}

\vspace{8pt}
\begin{tcbraster}[raster columns=2, raster equal height=rows, raster column skip=1em]
\begin{tcolorbox}[modelbox, title={GPT suggestion (incorrect)}, colback=sugwrong!8, colframe=sugwrong, colbacktitle=sugwrong]
\textbf{Primary diagnosis:} Bronchitis\\[3pt]
The chest radiograph shows relatively preserved bilateral lung lucency and no cardiac enlargement. No focal consolidation, pleural effusion, or pneumothorax is seen. Markings in both lower lungs and around the hila are mildly thickened. With cough and chest tightness, bronchitis or airway inflammation is favored.
\end{tcolorbox}
\begin{tcolorbox}[modelbox, title={Kimi suggestion (incorrect)}, colback=sugwrong!8, colframe=sugwrong, colbacktitle=sugwrong]
\textbf{Primary diagnosis:} Bronchitis\\[3pt]
The chest radiograph shows thickened and disorganized lung markings bilaterally and a small amount of patchy ill-defined opacity in the right lower lung. The cardiac and diaphragmatic contours are unremarkable. With cough and chest tightness, the findings are consistent with bronchitis, possibly accompanied by mild right lower-lung infection.
\end{tcolorbox}
\end{tcbraster}
\end{tcolorbox}

\clearpage
\subsection*{Supplementary Appendix 2. Prespecified Diagnostic Lexicon for Free-Text Response Scoring}

\textit{The} purpose of this appendix defines the diagnostic categories, acceptable diagnostic expressions, and exclusions used to score participants' free-text primary diagnoses before and after AI presentation. Each case was linked to its expert-established reference diagnosis before automated scoring.

\paragraph{Scoring Rules.}
\begin{enumerate}[leftmargin=1.4em]
\item Responses were considered correct when the primary diagnosis matched the reference diagnosis or an accepted diagnostic expression. Qualifiers such as ``likely'' or ``suggestive of'' did not invalidate an otherwise correct primary diagnosis.
\item Negated diagnoses were not accepted. Ambiguous statements, contradictory diagnoses, and unclear abbreviations required expert adjudication.
\item Scoring focused on the reference diagnostic category. Omitted laterality, site, or severity did not invalidate a correct category-level diagnosis. Explicitly incorrect modifiers required expert adjudication.
\item Automated matching used the locked lexicon. Two senior radiologists independently reviewed all classifications while masked to participant identity and study group. Disagreements were resolved by consensus.
\end{enumerate}

\paragraph{Diagnostic Category Lexicon.}
Abbreviations: AI = artificial intelligence; GPT = generative pretrained transformer.

\begin{table}[htbp]
\centering
% 加上下面这一行，让表题左对齐
\captionsetup{justification=raggedright, singlelinecheck=false}
\caption*{\textbf{Appendix Table A1.} Accepted Diagnostic Expressions and Examples Not Accepted as Standalone Answers}
\label{tab:a1}
\begin{threeparttable}
\setlength{\tabcolsep}{5pt}
\renewcommand{\arraystretch}{1.2}
\makebox[\textwidth][c]{\begin{tabular}{>{\raggedright\arraybackslash}p{0.20\linewidth}>{\raggedright\arraybackslash}p{0.26\linewidth}>{\raggedright\arraybackslash}p{0.26\linewidth}>{\raggedright\arraybackslash}p{0.22\linewidth}}
\toprule
Reference diagnostic category & Accepted diagnostic expressions in Chinese & English equivalents and diagnostic scope & Examples of descriptive findings or alternative diagnoses not accepted alone \\
\midrule
Pneumothorax / pleural air \zh{气胸/胸腔积气} &
\zh{气胸；左侧/右侧/双侧气胸；胸腔积气；张力性气胸（存在气胸时）} &
Pneumothorax; unilateral or bilateral pneumothorax; pleural air/gas; tension pneumothorax when pneumothorax is present &
\zh{肺大疱；胸膜增厚；胸腔积液；肺气肿} \\
\addlinespace
Pneumonia \zh{肺炎} &
\zh{肺炎；支气管肺炎；大叶性肺炎；肺部感染；感染性肺实变；明确诊断为炎性浸润/炎症} &
Pneumonia; bronchopneumonia; lobar pneumonia; pulmonary infection; infectious consolidation; inflammatory infiltrate stated as a diagnosis &
\zh{肺纹理增多；斑片影；肺实变或浸润影但未提示感染；支气管炎} \\
\addlinespace
Lung cancer \zh{肺癌} &
\zh{肺癌；原发性肺癌；支气管肺癌；中央型/周围型肺癌；肺恶性肿瘤；肺恶性占位；明确提示恶性的肺部肿块；肺癌病理亚型} &
Primary lung cancer; bronchogenic carcinoma; central/peripheral lung cancer; malignant pulmonary neoplasm or mass; a specified lung-cancer histologic subtype &
\zh{肺结节；肺肿块；肺占位（未提示恶性）；肺转移瘤；肺炎} \\
\addlinespace
Bronchitis \zh{支气管炎} &
\zh{支气管炎；急性/慢性支气管炎；气管支气管炎；支气管感染或炎症} &
Bronchitis; acute or chronic bronchitis; tracheobronchitis; bronchial infection/inflammation &
\zh{肺纹理增粗或增多；肺门影增浓；肺炎；慢性阻塞性肺疾病} \\
\addlinespace
Urinary tract calculi \zh{泌尿系结石} &
\zh{泌尿系结石；尿路结石；肾结石；输尿管结石；肾输尿管结石；肾区/输尿管区结石影（按病例参考部位复核）} &
Urolithiasis/urinary tract stone; nephrolithiasis/renal calculus; ureterolithiasis/ureteral calculus; renal and ureteral calculi &
\zh{盆腔钙化；静脉石；肾钙化；腹部钙化影（未明确泌尿系来源）} \\
\addlinespace
Bowel obstruction \zh{肠梗阻} &
\zh{肠梗阻；机械性肠梗阻；小肠/结肠梗阻；高位/低位肠梗阻；完全性/不完全性肠梗阻；绞窄性肠梗阻} &
Bowel/intestinal obstruction; mechanical obstruction; small- or large-bowel obstruction; complete, partial, high, low, or strangulated obstruction &
\zh{肠胀气；肠管扩张；气液平面；肠麻痹/麻痹性肠梗阻（除非参考诊断为该类别）} \\
\addlinespace
Hollow-viscus perforation / pneumoperitoneum \zh{空腔脏器穿孔/气腹} &
\zh{空腔脏器穿孔；消化道/胃肠道穿孔；胃/十二指肠/肠穿孔；气腹；腹腔游离气体；膈下游离气体；腹腔积气} &
Hollow-viscus or gastrointestinal perforation; gastric, duodenal, or intestinal perforation; pneumoperitoneum; free intraperitoneal or subdiaphragmatic air &
\zh{肠梗阻；腹胀；腹膜炎；腹腔积液；皮下气肿（未提示腹腔游离气体）} \\
\bottomrule
\end{tabular}}
\end{threeparttable}
\end{table}

\clearpage
% ----- Supplementary numerical tables -----
\subsection*{Supplementary Tables}
\renewcommand{\arraystretch}{1.2}
\setlength{\tabcolsep}{3.5pt}

\begin{table}[H]
\centering
% 加上下面这一行，让表题左对齐
\captionsetup{justification=raggedright, singlelinecheck=false}
\caption{Participant Characteristics by Specialty and Support Condition}
\label{tab:s1}
\begin{threeparttable}
\small
\setlength{\tabcolsep}{4pt}
\renewcommand{\arraystretch}{1.2}
\makebox[\textwidth][c]{\begin{adjustbox}{max width=\textwidth}
\begin{tabular}{lccccc}
\toprule
Characteristic & Group A & Group B & Group C & Statistic & $P$ value \\
\midrule
\multicolumn{6}{l}{\textit{Radiology residents}} \\
Number & 20 & 21 & 19 & --- & --- \\
Age (y) & $24.0 \pm 1.8$ & $24.0 \pm 1.6$ & $24.2 \pm 1.2$ & $F = 0.053$ & .949 \\
Sex, male & 5 (25.0\%) & 9 (42.9\%) & 8 (42.1\%) & $\chi^2 = 1.761$ & .415 \\
Experience (mo) & 9 (9--20) & 9 (9--20) & 20 (9--21) & $H = 0.919$ & .632 \\
\multicolumn{6}{l}{\textit{Non-radiology residents}} \\
Number & 20 & 21 & 22 & --- & --- \\
Age (y) & $24.4 \pm 1.5$ & $24.0 \pm 1.2$ & $23.8 \pm 1.3$ & $F = 1.095$ & .341 \\
Sex, male & 11 (55.0\%) & 13 (61.9\%) & 12 (54.5\%) & $\chi^2 = 0.293$ & .864 \\
Experience (mo) & 9 (9--19) & 19 (9--20) & 19 (9--20) & $H = 2.064$ & .356 \\
\bottomrule
\end{tabular}
\end{adjustbox}}
\end{threeparttable}
\fulltablenote{Note.---Data are mean $\pm$ SD, median (IQR), or $n$ (\%). Age was compared using one-way analysis of variance, experience using the Kruskal--Wallis test, and sex using the Pearson $\chi^2$ test.\par $F$ = analysis-of-variance statistic; $H$ = Kruskal--Wallis statistic; SD = standard deviation.}
\end{table}

\begin{table}[H]
\centering
% 加上下面这一行，让表题左对齐
\captionsetup{justification=raggedright, singlelinecheck=false}
\caption{Participant-Level Outcomes by Specialty}
\label{tab:s2}
\begin{threeparttable}
\small
\setlength{\tabcolsep}{3.5pt}
\makebox[\textwidth][c]{\begin{adjustbox}{max width=\textwidth}
\begin{tabular}{lccccc}
\toprule
Outcome & \makecell{Radiology ($n=60$)} & \makecell{Non-radiology ($n=63$)} & \makecell{Mean difference (95\% CI)} & \makecell{Welch $t$ (df)} & $P$ value \\
\midrule
AI-unassisted accuracy (\%) & $48.8 \pm 13.0$ & $43.9 \pm 9.0$ & 4.89 (0.88 to 8.90) & 2.42 (104.7) & .017 \\
AI-assisted accuracy (\%) & $63.2 \pm 9.3$ & $59.9 \pm 6.2$ & 3.28 (0.44 to 6.11) & 2.29 (102.2) & .024 \\
Accuracy change (pp) & $14.4 \pm 10.3$ & $16.0 \pm 7.2$ & $-1.62$ ($-4.81$ to 1.58) & $-1.00$ (104.9) & .318 \\
Total interpretation time (min) & $52.0 \pm 23.7$ & $55.1 \pm 27.3$ & $-3.07$ ($-12.18$ to 6.04) & $-0.67$ (120.0) & .505 \\
AI-assisted interpretation time (min) & $21.2 \pm 8.3$ & $21.5 \pm 8.9$ & $-0.25$ ($-3.31$ to 2.81) & $-0.16$ (120.9) & .871 \\
Confidence change (points) & $0.13 \pm 0.19$ & $0.24 \pm 0.27$ & $-0.12$ ($-0.20$ to $-0.04$) & $-2.85$ (113.3) & .005 \\
\bottomrule
\end{tabular}
\end{adjustbox}}
\end{threeparttable}
\fulltablenote{Note.---Group A: GPT; group B: GPT plus Kimi; group C: GPT plus Gemini. Radiology residents: $n = 60$ (A/B/C, 20/21/19); non-radiology residents: $n = 63$ (20/21/22).\par Values are mean $\pm$ SD. Differences are radiology minus non-radiology. Between-specialty $P$ values are unadjusted.\par CI = confidence interval; SD = standard deviation; df = degrees of freedom; pp = percentage points.}
\end{table}

\clearpage

\begin{table}[H]
\centering
% 加上下面这一行，让表题左对齐
\captionsetup{justification=raggedright, singlelinecheck=false}
\caption{Diagnostic Accuracy Before and After AI Assistance by Specialty and Support Condition}
\label{tab:s3}
\begin{threeparttable}
\small
\setlength{\tabcolsep}{3.5pt}
\makebox[\textwidth][c]{\begin{adjustbox}{max width=\textwidth}
\begin{tabular}{llcccccc}
\toprule
Population & Group & $n$ & \makecell{AI-unassisted (\%)} & \makecell{AI-assisted (\%)} & \makecell{Change, pp (95\% CI)} & \makecell{Paired $t$ (df)} & $P$ value \\
\midrule
All participants & A & 40 & $45.7 \pm 10.1$ & $59.5 \pm 7.4$ & 13.87 (11.27 to 16.48) & 10.77 (39) & $< .001$ \\
All participants & B & 42 & $47.9 \pm 12.8$ & $62.9 \pm 9.2$ & 15.00 (12.01 to 17.99) & 10.13 (41) & $< .001$ \\
All participants & C & 41 & $45.3 \pm 11.0$ & $62.2 \pm 6.9$ & 16.83 (14.10 to 19.56) & 12.45 (40) & $< .001$ \\
Radiology residents & A & 20 & $48.4 \pm 9.3$ & $58.0 \pm 8.4$ & 9.58 (6.26 to 12.91) & 6.03 (19) & $< .001$ \\
Radiology residents & B & 21 & $49.8 \pm 15.6$ & $66.1 \pm 10.8$ & 16.27 (11.42 to 21.12) & 7.00 (20) & $< .001$ \\
Radiology residents & C & 19 & $48.1 \pm 13.7$ & $65.5 \pm 5.5$ & 17.45 (11.98 to 22.93) & 6.70 (18) & $< .001$ \\
Non-radiology residents & A & 20 & $42.9 \pm 10.3$ & $61.1 \pm 6.0$ & 18.16 (14.95 to 21.37) & 11.84 (19) & $< .001$ \\
Non-radiology residents & B & 21 & $45.9 \pm 9.2$ & $59.6 \pm 6.0$ & 13.73 (9.86 to 17.60) & 7.41 (20) & $< .001$ \\
Non-radiology residents & C & 22 & $43.0 \pm 7.5$ & $59.2 \pm 6.7$ & 16.29 (13.80 to 18.77) & 13.62 (21) & $< .001$ \\
\bottomrule
\end{tabular}
\end{adjustbox}}
\end{threeparttable}
\fulltablenote{Note.---Values are mean $\pm$ SD. Change is final minus initial accuracy. Paired tests are two-sided and unadjusted.\par CI = confidence interval; SD = standard deviation; df = degrees of freedom; pp = percentage points.}
\end{table}

\begin{table}[H]
\centering
% 加上下面这一行，让表题左对齐
\captionsetup{justification=raggedright, singlelinecheck=false}
\caption{Accuracy Change, Interpretation Time, and Confidence Change by Specialty and Support Condition}
\label{tab:s4}
\begin{threeparttable}
\small
\setlength{\tabcolsep}{3.5pt}
\makebox[\textwidth][c]{\begin{adjustbox}{max width=\textwidth}
\begin{tabular}{llccccc}
\toprule
Population & Outcome & Group A & Group B & Group C & \makecell{Welch $F$ (df1, df2)} & $P$ value \\
\midrule
Radiology residents & Accuracy change (pp) & $9.6 \pm 7.1$ & $16.3 \pm 10.6$ & $17.5 \pm 11.4$ & 4.67 (2, 36.0) & .016 \\
Radiology residents & Interpretation time (min) & $59.0 \pm 30.1$ & $52.6 \pm 19.3$ & $44.0 \pm 18.7$ & 2.01 (2, 36.8) & .148 \\
Radiology residents & Confidence change (points) & $0.16 \pm 0.29$ & $0.09 \pm 0.10$ & $0.13 \pm 0.14$ & 1.00 (2, 33.2) & .380 \\
Non-radiology residents & Accuracy change (pp) & $18.2 \pm 6.9$ & $13.7 \pm 8.5$ & $16.3 \pm 5.6$ & 1.67 (2, 38.4) & .201 \\
Non-radiology residents & Interpretation time (min) & $57.6 \pm 27.4$ & $62.7 \pm 35.0$ & $45.5 \pm 13.8$ & 3.21 (2, 33.3) & .053 \\
Non-radiology residents & Confidence change (points) & $0.28 \pm 0.19$ & $0.22 \pm 0.20$ & $0.24 \pm 0.37$ & 0.41 (2, 38.8) & .666 \\
\bottomrule
\end{tabular}
\end{adjustbox}}
\end{threeparttable}
\fulltablenote{Note.---Values are mean $\pm$ SD. Omnibus $P$ values are unadjusted.\par CI = confidence interval; SD = standard deviation; df = degrees of freedom; pp = percentage points.}
\end{table}

\clearpage

\begin{table}[H]
\centering
\ContinuedFloat
% 加上下面这一行，让表题左对齐
\captionsetup{justification=raggedright, singlelinecheck=false}
\caption{Accuracy Change, Interpretation Time, and Confidence Change by Specialty and Support Condition (continued). Radiology residents; pairwise contrasts}
\label{tab:s4b}
\begin{threeparttable}
% ... (后面的代码保持不变)
\small
\setlength{\tabcolsep}{3.5pt}
\makebox[\textwidth][c]{\begin{adjustbox}{max width=\textwidth}
\begin{tabular}{llcccc}
\toprule
Outcome & Contrast & \makecell{Mean difference (95\% CI)} & \makecell{Welch $t$ (df)} & $P$ value & Holm $P$ \\
\midrule
Accuracy change (pp) & B--A & 6.69 (0.97 to 12.40) & 2.38 (35.0) & .023 & .030 \\
Accuracy change (pp) & C--A & 7.87 (1.64 to 14.11) & 2.58 (30.0) & .015 & .030 \\
Accuracy change (pp) & B--C & $-1.19$ ($-8.26$ to 5.89) & $-0.34$ (37.0) & .736 & NA \\
Interpretation time (min) & B--A & $-6.38$ ($-22.57$ to 9.80) & $-0.80$ (32.1) & .427 & .427 \\
Interpretation time (min) & C--A & $-14.96$ ($-31.23$ to 1.31) & $-1.87$ (32.0) & .070 & .140 \\
Interpretation time (min) & B--C & 8.58 ($-3.58$ to 20.73) & 1.43 (37.8) & .161 & NA \\
Confidence change (points) & B--A & $-0.08$ ($-0.22$ to 0.07) & $-1.10$ (23.2) & .282 & .565 \\
Confidence change (points) & C--A & $-0.03$ ($-0.18$ to 0.12) & $-0.45$ (28.0) & .656 & .656 \\
Confidence change (points) & B--C & $-0.04$ ($-0.12$ to 0.04) & $-1.08$ (31.7) & .287 & NA \\
\bottomrule
\end{tabular}
\end{adjustbox}}
\end{threeparttable}
\fulltablenote{Note.---Two-sided Welch $t$ tests. Differences are first-named minus second-named group. Holm adjustment was applied to the B--A and C--A comparisons within each outcome and specialty stratum. B--C is exploratory; its adjusted $P$ value is not applicable. CIs are pointwise.\par CI = confidence interval; df = degrees of freedom; pp = percentage points; NA = not applicable.}
\end{table}

\begin{table}[H]
\centering
\ContinuedFloat
% 加上下面这一行，让表题左对齐
\captionsetup{justification=raggedright, singlelinecheck=false}
\caption{Accuracy Change, Interpretation Time, and Confidence Change by Specialty and Support Condition (continued). Non-radiology residents; pairwise contrasts}
\label{tab:s4c}
\begin{threeparttable}
\small
\setlength{\tabcolsep}{3.5pt}
\makebox[\textwidth][c]{\begin{adjustbox}{max width=\textwidth}
\begin{tabular}{llcccc}
\toprule
Outcome & Contrast & \makecell{Mean difference (95\% CI)} & \makecell{Welch $t$ (df)} & $P$ value & Holm $P$ \\
\midrule
Accuracy change (pp) & B--A & $-4.44$ ($-9.31$ to 0.44) & $-1.84$ (38.0) & .073 & .146 \\
Accuracy change (pp) & C--A & $-1.88$ ($-5.82$ to 2.06) & $-0.97$ (36.8) & .340 & .340 \\
Accuracy change (pp) & B--C & $-2.56$ ($-7.04$ to 1.92) & $-1.16$ (34.4) & .254 & NA \\
Interpretation time (min) & B--A & 5.17 ($-14.66$ to 25.00) & 0.53 (37.6) & .601 & .601 \\
Interpretation time (min) & C--A & $-12.02$ ($-25.96$ to 1.92) & $-1.77$ (27.4) & .088 & .176 \\
Interpretation time (min) & B--C & 17.19 (0.37 to 34.01) & 2.10 (25.8) & .046 & NA \\
Confidence change (points) & B--A & $-0.06$ ($-0.18$ to 0.07) & $-0.90$ (39.0) & .373 & .747 \\
Confidence change (points) & C--A & $-0.04$ ($-0.22$ to 0.14) & $-0.45$ (32.2) & .656 & .747 \\
Confidence change (points) & B--C & $-0.01$ ($-0.20$ to 0.17) & $-0.17$ (32.8) & .869 & NA \\
\bottomrule
\end{tabular}
\end{adjustbox}}
\end{threeparttable}
\fulltablenote{Note.---Two-sided Welch $t$ tests. Differences are first-named minus second-named group. Holm adjustment was applied to the B--A and C--A comparisons within each outcome and specialty stratum. B--C is exploratory; its adjusted $P$ value is not applicable. CIs are pointwise.\par CI = confidence interval; df = degrees of freedom; pp = percentage points; NA = not applicable.}
\end{table}

\clearpage

\begin{table}[H]
\centering
% 加上下面这一行，让表题左对齐
\captionsetup{justification=raggedright, singlelinecheck=false}
\caption{Diagnostic Accuracy in Cases With a Correct Common GPT Suggestion}
\label{tab:s5}
\begin{threeparttable}
\small
\setlength{\tabcolsep}{3.5pt}
\makebox[\textwidth][c]{\begin{adjustbox}{max width=\textwidth}
\begin{tabular}{llccccc}
\toprule
Population & Outcome & Group A & Group B & Group C & \makecell{Welch $F$ (df1, df2)} & $P$ \\
\midrule
Radiology residents & AI-unassisted accuracy (\%) & $57.6 \pm 9.1$ & $57.6 \pm 16.5$ & $53.3 \pm 15.8$ & 0.55 (2, 34.8) & .580 \\
Radiology residents & AI-assisted accuracy (\%) & $78.5 \pm 12.5$ & $80.1 \pm 9.8$ & $79.1 \pm 8.6$ & 0.12 (2, 37.4) & .889 \\
Radiology residents & Accuracy change (pp) & $20.9 \pm 11.7$ & $22.5 \pm 15.1$ & $25.8 \pm 14.3$ & 0.67 (2, 37.4) & .517 \\
Non-radiology residents & AI-unassisted accuracy (\%) & $52.3 \pm 10.7$ & $54.1 \pm 11.4$ & $48.5 \pm 8.8$ & 1.79 (2, 38.9) & .180 \\
Non-radiology residents & AI-assisted accuracy (\%) & $87.4 \pm 7.2$ & $74.8 \pm 10.0$ & $74.5 \pm 9.3$ & 16.92 (2, 39.5) & $< .001$ \\
Non-radiology residents & Accuracy change (pp) & $35.1 \pm 10.2$ & $20.8 \pm 10.8$ & $26.0 \pm 8.8$ & 9.88 (2, 39.2) & $< .001$ \\
\bottomrule
\end{tabular}
\end{adjustbox}}
\end{threeparttable}
\fulltablenote{Note.---Values are mean $\pm$ SD. Omnibus $P$ values are unadjusted. Conditional accuracy uses the common 39 GPT-correct cases as the denominator.\par CI = confidence interval; SD = standard deviation; df = degrees of freedom; pp = percentage points.}
\end{table}

\begin{table}[H]
\centering
% 加上下面这一行，让表题左对齐
\captionsetup{justification=raggedright, singlelinecheck=false}
\caption{Diagnostic Accuracy in Cases With an Incorrect Common GPT Suggestion}
\label{tab:s6}
\begin{threeparttable}
\small
\setlength{\tabcolsep}{3.5pt}
\makebox[\textwidth][c]{\begin{adjustbox}{max width=\textwidth}
\begin{tabular}{llccccc}
\toprule
Population & Outcome & Group A & Group B & Group C & \makecell{Welch $F$ (df1, df2)} & $P$ \\
\midrule
Radiology residents & AI-unassisted accuracy (\%) & $31.4 \pm 14.9$ & $35.4 \pm 17.6$ & $38.3 \pm 14.3$ & 1.08 (2, 38.0) & .349 \\
Radiology residents & AI-assisted accuracy (\%) & $20.0 \pm 13.5$ & $40.1 \pm 17.9$ & $40.4 \pm 9.4$ & 15.90 (2, 36.6) & $< .001$ \\
Radiology residents & Accuracy change (pp) & $-11.4 \pm 10.6$ & $4.8 \pm 10.1$ & $2.0 \pm 9.7$ & 13.70 (2, 37.9) & $< .001$ \\
Non-radiology residents & AI-unassisted accuracy (\%) & $25.5 \pm 13.5$ & $30.6 \pm 12.6$ & $32.7 \pm 9.8$ & 1.90 (2, 38.4) & .163 \\
Non-radiology residents & AI-assisted accuracy (\%) & $12.1 \pm 12.9$ & $31.3 \pm 11.7$ & $31.0 \pm 11.1$ & 15.66 (2, 39.4) & $< .001$ \\
Non-radiology residents & Accuracy change (pp) & $-13.3 \pm 9.1$ & $0.7 \pm 9.3$ & $-1.7 \pm 9.6$ & 13.35 (2, 39.9) & $< .001$ \\
\bottomrule
\end{tabular}
\end{adjustbox}}
\end{threeparttable}
\fulltablenote{Note.---Values are mean $\pm$ SD. Omnibus $P$ values are unadjusted. Conditional accuracy uses the common 21 GPT-incorrect cases as the denominator.\par CI = confidence interval; SD = standard deviation; df = degrees of freedom; pp = percentage points.}
\end{table}

\clearpage

\begin{table}[H]
\centering
% 加上下面这一行，让表题左对齐
\captionsetup{justification=raggedright, singlelinecheck=false}
\caption{Pairwise Accuracy Contrasts by Specialty and GPT Suggestion Correctness. Radiology residents; GPT correct (39 cases); pairwise contrasts}
\label{tab:s7}
\label{tab:s7a}
\begin{threeparttable}
\small
\setlength{\tabcolsep}{3.5pt}
\makebox[\textwidth][c]{\begin{adjustbox}{max width=\textwidth}
\begin{tabular}{llcccc}
\toprule
Outcome & Contrast & \makecell{Mean difference (95\% CI)} & \makecell{Welch $t$ (df)} & $P$ value & Holm $P$ \\
\midrule
AI-unassisted accuracy (pp) & B--A & 0.07 ($-8.33$ to 8.47) & 0.02 (31.4) & .987 & .987 \\
AI-unassisted accuracy (pp) & C--A & $-4.26$ ($-12.75$ to 4.23) & $-1.03$ (28.4) & .313 & .626 \\
AI-unassisted accuracy (pp) & B--C & 4.32 ($-6.00$ to 14.64) & 0.85 (37.9) & .401 & NA \\
AI-assisted accuracy (pp) & B--A & 1.64 ($-5.50$ to 8.78) & 0.46 (36.1) & .645 & 1.000 \\
AI-assisted accuracy (pp) & C--A & 0.62 ($-6.33$ to 7.57) & 0.18 (33.8) & .857 & 1.000 \\
AI-assisted accuracy (pp) & B--C & 1.02 ($-4.88$ to 6.92) & 0.35 (38.0) & .729 & NA \\
Accuracy change (pp) & B--A & 1.57 ($-6.95$ to 10.09) & 0.37 (37.4) & .711 & .711 \\
Accuracy change (pp) & C--A & 4.88 ($-3.62$ to 13.37) & 1.17 (34.8) & .251 & .503 \\
Accuracy change (pp) & B--C & $-3.31$ ($-12.72$ to 6.10) & $-0.71$ (37.9) & .481 & NA \\
\bottomrule
\end{tabular}
\end{adjustbox}}
\end{threeparttable}
\fulltablenote{Note.---Two-sided Welch $t$ tests. Differences are first-named minus second-named group. Holm adjustment was applied to the B--A and C--A comparisons within each outcome, specialty stratum, and GPT-correctness subset. B--C is exploratory; its adjusted $P$ value is not applicable. CIs are pointwise.\par CI = confidence interval; df = degrees of freedom; pp = percentage points; NA = not applicable.}
\end{table}

\begin{table}[H]
\centering
\ContinuedFloat
% 加上下面这一行，让表题左对齐
\captionsetup{justification=raggedright, singlelinecheck=false}
\caption{Pairwise Accuracy Contrasts by Specialty and GPT Suggestion Correctness (continued). Non-radiology residents; GPT correct (39 cases); pairwise contrasts}
\label{tab:s7b}
\begin{threeparttable}
\small
\setlength{\tabcolsep}{3.5pt}
\makebox[\textwidth][c]{\begin{adjustbox}{max width=\textwidth}
\begin{tabular}{llcccc}
\toprule
Outcome & Contrast & \makecell{Mean difference (95\% CI)} & \makecell{Welch $t$ (df)} & $P$ value & Holm $P$ \\
\midrule
AI-unassisted accuracy (pp) & B--A & 1.78 ($-5.21$ to 8.78) & 0.52 (39.0) & .609 & .609 \\
AI-unassisted accuracy (pp) & C--A & $-3.82$ ($-9.99$ to 2.34) & $-1.26$ (36.8) & .217 & .434 \\
AI-unassisted accuracy (pp) & B--C & 5.61 ($-0.70$ to 11.91) & 1.80 (37.6) & .080 & NA \\
AI-assisted accuracy (pp) & B--A & $-12.59$ ($-18.10$ to $-7.07$) & $-4.63$ (36.3) & $< .001$ & $< .001$ \\
AI-assisted accuracy (pp) & C--A & $-12.96$ ($-18.13$ to $-7.79$) & $-5.07$ (39.0) & $< .001$ & $< .001$ \\
AI-assisted accuracy (pp) & B--C & 0.37 ($-5.61$ to 6.35) & 0.13 (40.4) & .901 & NA \\
Accuracy change (pp) & B--A & $-14.37$ ($-20.99$ to $-7.76$) & $-4.39$ (39.0) & $< .001$ & $< .001$ \\
Accuracy change (pp) & C--A & $-9.14$ ($-15.09$ to $-3.19$) & $-3.11$ (37.7) & .004 & .004 \\
Accuracy change (pp) & B--C & $-5.23$ ($-11.31$ to 0.84) & $-1.74$ (38.5) & .089 & NA \\
\bottomrule
\end{tabular}
\end{adjustbox}}
\end{threeparttable}
\fulltablenote{Note.---Two-sided Welch $t$ tests. Differences are first-named minus second-named group. Holm adjustment was applied to the B--A and C--A comparisons within each outcome, specialty stratum, and GPT-correctness subset. B--C is exploratory; its adjusted $P$ value is not applicable. CIs are pointwise.\par CI = confidence interval; df = degrees of freedom; pp = percentage points; NA = not applicable.}
\end{table}

\clearpage

\begin{table}[H]
\centering
\ContinuedFloat
% 加上下面这一行，让表题左对齐
\captionsetup{justification=raggedright, singlelinecheck=false}
\caption{Pairwise Accuracy Contrasts by Specialty and GPT Suggestion Correctness (continued). Radiology residents; GPT incorrect (21 cases); pairwise contrasts}
\label{tab:s7c}
\begin{threeparttable}
\small
\setlength{\tabcolsep}{3.5pt}
\makebox[\textwidth][c]{\begin{adjustbox}{max width=\textwidth}
\begin{tabular}{llcccc}
\toprule
Outcome & Contrast & \makecell{Mean difference (95\% CI)} & \makecell{Welch $t$ (df)} & $P$ value & Holm $P$ \\
\midrule
AI-unassisted accuracy (pp) & B--A & 3.95 ($-6.33$ to 14.22) & 0.78 (38.5) & .442 & .442 \\
AI-unassisted accuracy (pp) & C--A & 6.92 ($-2.55$ to 16.38) & 1.48 (37.0) & .147 & .294 \\
AI-unassisted accuracy (pp) & B--C & $-2.97$ ($-13.21$ to 7.26) & $-0.59$ (37.6) & .560 & NA \\
AI-assisted accuracy (pp) & B--A & 20.14 (10.15 to 30.13) & 4.08 (37.2) & $< .001$ & $< .001$ \\
AI-assisted accuracy (pp) & C--A & 20.35 (12.79 to 27.91) & 5.47 (34.0) & $< .001$ & $< .001$ \\
AI-assisted accuracy (pp) & B--C & $-0.21$ ($-9.31$ to 8.88) & $-0.05$ (31.0) & .962 & NA \\
Accuracy change (pp) & B--A & 16.19 (9.63 to 22.75) & 4.99 (38.6) & $< .001$ & $< .001$ \\
Accuracy change (pp) & C--A & 13.43 (6.83 to 20.03) & 4.13 (36.9) & $< .001$ & $< .001$ \\
Accuracy change (pp) & B--C & 2.76 ($-3.58$ to 9.10) & 0.88 (37.9) & .384 & NA \\
\bottomrule
\end{tabular}
\end{adjustbox}}
\end{threeparttable}
\fulltablenote{Note.---Two-sided Welch $t$ tests. Differences are first-named minus second-named group. Holm adjustment was applied to the B--A and C--A comparisons within each outcome, specialty stratum, and GPT-correctness subset. B--C is exploratory; its adjusted $P$ value is not applicable. CIs are pointwise.\par CI = confidence interval; df = degrees of freedom; pp = percentage points; NA = not applicable.}
\end{table}

\begin{table}[H]
\centering
\ContinuedFloat
% 加上下面这一行，让表题左对齐
\captionsetup{justification=raggedright, singlelinecheck=false}
\caption{Pairwise Accuracy Contrasts by Specialty and GPT Suggestion Correctness (continued). Non-radiology residents; GPT incorrect (21 cases); pairwise contrasts}
\label{tab:s7d}
\begin{threeparttable}
\small
\setlength{\tabcolsep}{3.5pt}
\makebox[\textwidth][c]{\begin{adjustbox}{max width=\textwidth}
\begin{tabular}{llcccc}
\toprule
Outcome & Contrast & \makecell{Mean difference (95\% CI)} & \makecell{Welch $t$ (df)} & $P$ value & Holm $P$ \\
\midrule
AI-unassisted accuracy (pp) & B--A & 5.14 ($-3.11$ to 13.38) & 1.26 (38.4) & .215 & .215 \\
AI-unassisted accuracy (pp) & C--A & 7.21 ($-0.24$ to 14.66) & 1.97 (34.4) & .058 & .115 \\
AI-unassisted accuracy (pp) & B--C & $-2.07$ ($-9.04$ to 4.90) & $-0.60$ (37.8) & .551 & NA \\
AI-assisted accuracy (pp) & B--A & 19.15 (11.35 to 26.94) & 4.97 (38.2) & $< .001$ & $< .001$ \\
AI-assisted accuracy (pp) & C--A & 18.81 (11.27 to 26.35) & 5.05 (37.7) & $< .001$ & $< .001$ \\
AI-assisted accuracy (pp) & B--C & 0.34 ($-6.69$ to 7.37) & 0.10 (40.6) & .923 & NA \\
Accuracy change (pp) & B--A & 14.01 (8.20 to 19.83) & 4.87 (39.0) & $< .001$ & $< .001$ \\
Accuracy change (pp) & C--A & 11.60 (5.78 to 17.43) & 4.03 (39.9) & $< .001$ & $< .001$ \\
Accuracy change (pp) & B--C & 2.41 ($-3.41$ to 8.23) & 0.84 (41.0) & .408 & NA \\
\bottomrule
\end{tabular}
\end{adjustbox}}
\end{threeparttable}
\fulltablenote{Note.---Two-sided Welch $t$ tests. Differences are first-named minus second-named group. Holm adjustment was applied to the B--A and C--A comparisons within each outcome, specialty stratum, and GPT-correctness subset. B--C is exploratory; its adjusted $P$ value is not applicable. CIs are pointwise.\par CI = confidence interval; df = degrees of freedom; pp = percentage points; NA = not applicable.}
\end{table}

\clearpage

\begin{table}[H]
\centering
% 加上下面这一行，让表题左对齐
\captionsetup{justification=raggedright, singlelinecheck=false}
\caption{Diagnostic Revision Patterns by Specialty and Support Condition}
\label{tab:s8}
\begin{threeparttable}
\small
\setlength{\tabcolsep}{3.5pt}
\makebox[\textwidth][c]{\begin{adjustbox}{max width=\textwidth}
\begin{tabular}{llccccc}
\toprule
Population & Outcome & Group A & Group B & Group C & \makecell{Welch $F$ (df1, df2)} & $P$ \\
\midrule
Radiology residents & Incorrect $\rightarrow$ correct & $14.8 \pm 7.8$ & $19.9 \pm 11.0$ & $22.8 \pm 10.5$ & 3.98 (2, 36.9) & .027 \\
Radiology residents & Correct $\rightarrow$ incorrect & $5.2 \pm 3.9$ & $3.7 \pm 2.4$ & $5.4 \pm 3.3$ & 2.15 (2, 35.6) & .131 \\
Radiology residents & Incorrect $\rightarrow$ incorrect & $36.8 \pm 11.1$ & $30.2 \pm 11.2$ & $29.1 \pm 6.5$ & 3.55 (2, 36.1) & .039 \\
Radiology residents & Correct $\rightarrow$ correct & $43.3 \pm 7.8$ & $46.2 \pm 15.8$ & $42.7 \pm 12.9$ & 0.34 (2, 34.6) & .713 \\
Non-radiology residents & Incorrect $\rightarrow$ correct & $24.0 \pm 6.8$ & $19.7 \pm 8.4$ & $22.8 \pm 6.1$ & 1.66 (2, 38.9) & .204 \\
Non-radiology residents & Correct $\rightarrow$ incorrect & $5.8 \pm 3.9$ & $6.0 \pm 3.9$ & $6.5 \pm 3.2$ & 0.23 (2, 39.1) & .796 \\
Non-radiology residents & Incorrect $\rightarrow$ incorrect & $33.1 \pm 7.8$ & $34.4 \pm 6.1$ & $34.2 \pm 8.3$ & 0.20 (2, 39.2) & .821 \\
Non-radiology residents & Correct $\rightarrow$ correct & $37.1 \pm 9.8$ & $39.9 \pm 9.9$ & $36.4 \pm 6.9$ & 0.89 (2, 37.9) & .418 \\
\bottomrule
\end{tabular}
\end{adjustbox}}
\end{threeparttable}
\fulltablenote{Note.---Values are mean $\pm$ SD. Omnibus $P$ values are unadjusted. Transition values are percentages using all 60 cases as the denominator. Beneficial revision: incorrect $\rightarrow$ correct; harmful revision: correct $\rightarrow$ incorrect.\par CI = confidence interval; SD = standard deviation; df = degrees of freedom; pp = percentage points.}
\end{table}

\begin{table}[H]
\centering
\ContinuedFloat
% 加上下面这一行，让表题左对齐
\captionsetup{justification=raggedright, singlelinecheck=false}
\caption{Diagnostic Revision Patterns by Specialty and Support Condition (continued). Radiology residents; pairwise contrasts}
\label{tab:s8b}
\begin{threeparttable}
\small
\setlength{\tabcolsep}{3.5pt}
\makebox[\textwidth][c]{\begin{adjustbox}{max width=\textwidth}
\begin{tabular}{llcccc}
\toprule
Outcome & Contrast & \makecell{Mean difference (95\% CI)} & \makecell{Welch $t$ (df)} & $P$ value & Holm $P$ \\
\midrule
Incorrect $\rightarrow$ correct & B--A & 5.17 ($-0.84$ to 11.18) & 1.75 (36.1) & .089 & .089 \\
Incorrect $\rightarrow$ correct & C--A & 8.06 (2.01 to 14.10) & 2.71 (33.1) & .011 & .021 \\
Incorrect $\rightarrow$ correct & B--C & $-2.89$ ($-9.77$ to 3.99) & $-0.85$ (37.9) & .401 & NA \\
Correct $\rightarrow$ incorrect & B--A & $-1.52$ ($-3.60$ to 0.57) & $-1.48$ (31.1) & .148 & .297 \\
Correct $\rightarrow$ incorrect & C--A & 0.18 ($-2.17$ to 2.54) & 0.16 (36.5) & .875 & .875 \\
Correct $\rightarrow$ incorrect & B--C & $-1.70$ ($-3.58$ to 0.18) & $-1.84$ (32.5) & .074 & NA \\
Incorrect $\rightarrow$ incorrect & B--A & $-6.60$ ($-13.64$ to 0.45) & $-1.89$ (38.9) & .066 & .066 \\
Incorrect $\rightarrow$ incorrect & C--A & $-7.71$ ($-13.61$ to $-1.82$) & $-2.67$ (30.9) & .012 & .024 \\
Incorrect $\rightarrow$ incorrect & B--C & 1.12 ($-4.72$ to 6.95) & 0.39 (32.6) & .700 & NA \\
Correct $\rightarrow$ correct & B--A & 2.94 ($-4.93$ to 10.81) & 0.76 (29.5) & .451 & .902 \\
Correct $\rightarrow$ correct & C--A & $-0.53$ ($-7.56$ to 6.50) & $-0.15$ (29.1) & .878 & .902 \\
Correct $\rightarrow$ correct & B--C & 3.47 ($-5.73$ to 12.67) & 0.76 (37.7) & .450 & NA \\
\bottomrule
\end{tabular}
\end{adjustbox}}
\end{threeparttable}
\fulltablenote{Note.---Two-sided Welch $t$ tests. Differences are first-named minus second-named group. Holm adjustment was applied to the B--A and C--A comparisons within each revision outcome and specialty stratum. B--C is exploratory; its adjusted $P$ value is not applicable. CIs are pointwise. Beneficial revision: incorrect $\rightarrow$ correct; harmful revision: correct $\rightarrow$ incorrect.\par CI = confidence interval; df = degrees of freedom; pp = percentage points; NA = not applicable.}
\end{table}

\clearpage

\begin{table}[H]
\centering
\ContinuedFloat
% 加上下面这一行，让表题左对齐
\captionsetup{justification=raggedright, singlelinecheck=false}
\caption{Diagnostic Revision Patterns by Specialty and Support Condition (continued). Non-radiology residents; pairwise contrasts}
\label{tab:s8c}
\begin{threeparttable}
\small
\setlength{\tabcolsep}{3.5pt}
\makebox[\textwidth][c]{\begin{adjustbox}{max width=\textwidth}
\begin{tabular}{llcccc}
\toprule
Outcome & Contrast & \makecell{Mean difference (95\% CI)} & \makecell{Welch $t$ (df)} & $P$ value & Holm $P$ \\
\midrule
Incorrect $\rightarrow$ correct & B--A & $-4.32$ ($-9.16$ to 0.52) & $-1.81$ (38.0) & .079 & .157 \\
Incorrect $\rightarrow$ correct & C--A & $-1.20$ ($-5.24$ to 2.85) & $-0.60$ (38.2) & .553 & .553 \\
Incorrect $\rightarrow$ correct & B--C & $-3.12$ ($-7.68$ to 1.43) & $-1.39$ (36.2) & .173 & NA \\
Correct $\rightarrow$ incorrect & B--A & 0.12 ($-2.35$ to 2.59) & 0.10 (38.9) & .923 & 1.000 \\
Correct $\rightarrow$ incorrect & C--A & 0.68 ($-1.57$ to 2.93) & 0.61 (36.8) & .543 & 1.000 \\
Correct $\rightarrow$ incorrect & B--C & $-0.56$ ($-2.77$ to 1.64) & $-0.52$ (38.8) & .609 & NA \\
Incorrect $\rightarrow$ incorrect & B--A & 1.36 ($-3.10$ to 5.82) & 0.62 (36.0) & .540 & 1.000 \\
Incorrect $\rightarrow$ incorrect & C--A & 1.16 ($-3.87$ to 6.18) & 0.47 (39.9) & .644 & 1.000 \\
Incorrect $\rightarrow$ incorrect & B--C & 0.20 ($-4.28$ to 4.68) & 0.09 (38.6) & .928 & NA \\
Correct $\rightarrow$ correct & B--A & 2.84 ($-3.38$ to 9.06) & 0.92 (38.9) & .362 & .724 \\
Correct $\rightarrow$ correct & C--A & $-0.64$ ($-6.03$ to 4.74) & $-0.24$ (33.8) & .809 & .809 \\
Correct $\rightarrow$ correct & B--C & 3.48 ($-1.81$ to 8.77) & 1.33 (35.7) & .190 & NA \\
\bottomrule
\end{tabular}
\end{adjustbox}}
\end{threeparttable}
\fulltablenote{Note.---Two-sided Welch $t$ tests. Differences are first-named minus second-named group. Holm adjustment was applied to the B--A and C--A comparisons within each revision outcome and specialty stratum. B--C is exploratory; its adjusted $P$ value is not applicable. CIs are pointwise. Beneficial revision: incorrect $\rightarrow$ correct; harmful revision: correct $\rightarrow$ incorrect.\par CI = confidence interval; df = degrees of freedom; pp = percentage points; NA = not applicable.}
\end{table}

\begin{table}[H]
\centering
% 加上下面这一行，让表题左对齐
\captionsetup{justification=raggedright, singlelinecheck=false}
\caption{Dual-Suggestion Effects and Specialty Interactions. Accuracy change}
\label{tab:s9}
\label{tab:s9a}
\begin{threeparttable}
\small
\setlength{\tabcolsep}{3.5pt}
\makebox[\textwidth][c]{\begin{adjustbox}{max width=\textwidth}
\begin{tabular}{llcccc}
\toprule
Analysis & Contrast & \makecell{Effect estimate, pp (95\% CI)} & \makecell{Wald statistic} & $P$ value & Holm $P$ \\
\midrule
Radiology residents & Dual vs single & 7.28 (2.53 to 12.03) & $z = 3.00$ & .003 & NA \\
Radiology residents & B--A & 6.69 (1.03 to 12.34) & $z = 2.32$ & .021 & .024 \\
Radiology residents & C--A & 7.87 (1.73 to 14.02) & $z = 2.51$ & .012 & .024 \\
Non-radiology residents & Dual vs single & $-3.16$ ($-6.95$ to 0.64) & $z = -1.63$ & .103 & NA \\
Non-radiology residents & B--A & $-4.44$ ($-9.27$ to 0.40) & $z = -1.80$ & .072 & .144 \\
Non-radiology residents & C--A & $-1.88$ ($-5.79$ to 2.03) & $z = -0.94$ & .346 & .346 \\
Specialty interaction & Dual vs single & 10.44 (4.36 to 16.52) & $z = 3.36$ & $< .001$ & NA \\
Specialty interaction & B--A & 11.12 (3.68 to 18.56) & $z = 2.93$ & .003 & .007 \\
Specialty interaction & C--A & 9.75 (2.47 to 17.03) & $z = 2.62$ & .009 & .009 \\
Overall interaction & Group $\times$ specialty & NA & $\chi^2(2) = 11.38$ & .003 & NA \\
\bottomrule
\end{tabular}
\end{adjustbox}}
\end{threeparttable}
\fulltablenote{Note.---HC3 linear models. Specialty interaction is the radiology contrast minus the non-radiology contrast. Holm adjustment covers B--A/C--A within each outcome and analysis block. The dual-suggestion effect was defined as the equally weighted contrast $[(B + C)/2] - A$. Averaged contrasts and omnibus tests are unadjusted; NA = not applicable.\par CI = confidence interval; pp = percentage points.}
\end{table}

\clearpage

\begin{table}[H]
\centering
\ContinuedFloat
% 加上下面这一行，让表题左对齐
\captionsetup{justification=raggedright, singlelinecheck=false}
\caption{Dual-Suggestion Effects and Specialty Interactions (continued). Beneficial revisions}
\label{tab:s9b}
\begin{threeparttable}
\small
\setlength{\tabcolsep}{3.5pt}
\makebox[\textwidth][c]{\begin{adjustbox}{max width=\textwidth}
\begin{tabular}{llcccc}
\toprule
Analysis & Contrast & \makecell{Effect estimate, pp (95\% CI)} & \makecell{Wald statistic} & $P$ value & Holm $P$ \\
\midrule
Radiology residents & Dual vs single & 6.61 (1.72 to 11.51) & $z = 2.65$ & .008 & NA \\
Radiology residents & B--A & 5.17 ($-0.78$ to 11.12) & $z = 1.70$ & .089 & .089 \\
Radiology residents & C--A & 8.06 (2.08 to 14.04) & $z = 2.64$ & .008 & .017 \\
Non-radiology residents & Dual vs single & $-2.76$ ($-6.57$ to 1.05) & $z = -1.42$ & .156 & NA \\
Non-radiology residents & B--A & $-4.32$ ($-9.12$ to 0.48) & $z = -1.76$ & .078 & .156 \\
Non-radiology residents & C--A & $-1.20$ ($-5.21$ to 2.82) & $z = -0.58$ & .559 & .559 \\
Specialty interaction & Dual vs single & 9.37 (3.17 to 15.57) & $z = 2.96$ & .003 & NA \\
Specialty interaction & B--A & 9.49 (1.84 to 17.14) & $z = 2.43$ & .015 & .024 \\
Specialty interaction & C--A & 9.25 (2.05 to 16.46) & $z = 2.52$ & .012 & .024 \\
Overall interaction & Group $\times$ specialty & NA & $\chi^2(2) = 8.80$ & .012 & NA \\
\bottomrule
\end{tabular}
\end{adjustbox}}
\end{threeparttable}
\fulltablenote{Note.---HC3 linear models. Specialty interaction is the radiology contrast minus the non-radiology contrast. Holm adjustment covers B--A/C--A within each outcome and analysis block. The dual-suggestion effect was defined as the equally weighted contrast $[(B + C)/2] - A$. Averaged contrasts and omnibus tests are unadjusted; NA = not applicable.\par CI = confidence interval; pp = percentage points.}
\end{table}

\begin{table}[H]
\centering
\ContinuedFloat
% 加上下面这一行，让表题左对齐
\captionsetup{justification=raggedright, singlelinecheck=false}
\caption{Dual-Suggestion Effects and Specialty Interactions (continued). Harmful revisions}
\label{tab:s9c}
\begin{threeparttable}
\small
\setlength{\tabcolsep}{3.5pt}
\makebox[\textwidth][c]{\begin{adjustbox}{max width=\textwidth}
\begin{tabular}{llcccc}
\toprule
Analysis & Contrast & \makecell{Effect estimate, pp (95\% CI)} & \makecell{Wald statistic} & $P$ value & Holm $P$ \\
\midrule
Radiology residents & Dual vs single & $-0.67$ ($-2.66$ to 1.33) & $z = -0.65$ & .513 & NA \\
Radiology residents & B--A & $-1.52$ ($-3.57$ to 0.54) & $z = -1.45$ & .148 & .297 \\
Radiology residents & C--A & 0.18 ($-2.15$ to 2.52) & $z = 0.15$ & .877 & .877 \\
Non-radiology residents & Dual vs single & 0.40 ($-1.67$ to 2.47) & $z = 0.38$ & .705 & NA \\
Non-radiology residents & B--A & 0.12 ($-2.33$ to 2.57) & $z = 0.10$ & .924 & 1.000 \\
Non-radiology residents & C--A & 0.68 ($-1.55$ to 2.91) & $z = 0.60$ & .550 & 1.000 \\
Specialty interaction & Dual vs single & $-1.07$ ($-3.95$ to 1.81) & $z = -0.73$ & .468 & NA \\
Specialty interaction & B--A & $-1.63$ ($-4.84$ to 1.57) & $z = -1.00$ & .317 & .633 \\
Specialty interaction & C--A & $-0.50$ ($-3.73$ to 2.74) & $z = -0.30$ & .763 & .763 \\
Overall interaction & Group $\times$ specialty & NA & $\chi^2(2) = 1.14$ & .564 & NA \\
\bottomrule
\end{tabular}
\end{adjustbox}}
\end{threeparttable}
\fulltablenote{Note.---HC3 linear models. Specialty interaction is the radiology contrast minus the non-radiology contrast. Holm adjustment covers B--A/C--A within each outcome and analysis block. The dual-suggestion effect was defined as the equally weighted contrast $[(B + C)/2] - A$. Averaged contrasts and omnibus tests are unadjusted; NA = not applicable.\par CI = confidence interval; pp = percentage points.}
\end{table}

\clearpage

\begin{table}[H]
\centering
% 加上下面这一行，让表题左对齐
\captionsetup{justification=raggedright, singlelinecheck=false}
\caption{Revision Components by Shared GPT Suggestion Correctness, Specialty, and Support Condition}
\label{tab:s10}
\begin{threeparttable}
\small
\setlength{\tabcolsep}{3.5pt}
\makebox[\textwidth][c]{\begin{adjustbox}{max width=\textwidth}
\begin{tabular}{llcccccc}
\toprule
\makecell{Shared GPT suggestion} & \makecell{Revision component} & Group A & Group B & Group C & Overall $P$ & \makecell{B--A, Holm $P$} & \makecell{C--A, Holm $P$} \\
\midrule
\multicolumn{8}{l}{\textit{Radiology residents}} \\
Correct ($n = 39$) & Incorrect $\rightarrow$ correct & $22.1 \pm 11.6$ & $24.3 \pm 15.4$ & $29.0 \pm 14.0$ & .260 & .600 & .203 \\
& Correct $\rightarrow$ incorrect & $1.2 \pm 1.9$ & $1.8 \pm 2.0$ & $3.2 \pm 3.2$ & .063 & .279 & .040 \\
Incorrect ($n = 21$) & Incorrect $\rightarrow$ correct & $1.2 \pm 2.1$ & $11.8 \pm 7.8$ & $11.3 \pm 5.5$ & $< .001$ & $< .001$ & $< .001$ \\
& Correct $\rightarrow$ incorrect & $12.6 \pm 10.5$ & $7.0 \pm 5.8$ & $9.3 \pm 6.8$ & .116 & .089 & .245 \\
\multicolumn{8}{l}{\textit{Non-radiology residents}} \\
Correct ($n = 39$) & Incorrect $\rightarrow$ correct & $36.0 \pm 10.5$ & $25.3 \pm 11.8$ & $29.1 \pm 8.3$ & .012 & .007 & .025 \\
& Correct $\rightarrow$ incorrect & $0.9 \pm 1.7$ & $4.5 \pm 3.5$ & $3.1 \pm 3.4$ & $< .001$ & $< .001$ & .009 \\
Incorrect ($n = 21$) & Incorrect $\rightarrow$ correct & $1.7 \pm 2.3$ & $9.3 \pm 6.5$ & $11.0 \pm 6.6$ & $< .001$ & $< .001$ & $< .001$ \\
& Correct $\rightarrow$ incorrect & $15.0 \pm 9.3$ & $8.6 \pm 7.0$ & $12.8 \pm 7.5$ & .042 & .037 & .402 \\
\bottomrule
\end{tabular}
\end{adjustbox}}
\end{threeparttable}
\fulltablenote{Note.---Data are mean $\pm$ SD and represent the percentage of cases within each correctness stratum. Beneficial revisions are incorrect-to-correct transitions; harmful revisions are correct-to-incorrect transitions.\par Overall $P$ values were calculated using Welch analysis of variance. Pairwise $P$ values compare groups B and C separately with group A and were Holm-adjusted across the two prespecified comparisons within each specialty and revision component. B--C comparisons are not shown.\par The correctness strata were defined by the GPT suggestion shared across all three support conditions; correctness of the second suggestion in groups B and C could differ within each stratum.}
\end{table}

\clearpage

\begin{table}[H]
\centering
\captionsetup{justification=raggedright, singlelinecheck=false}
\caption{Outcomes in Cases for Which Both AI Models Were Incorrect}
\label{tab:s11}
\begin{threeparttable}
\small
\setlength{\tabcolsep}{3.5pt}
\makebox[\textwidth][c]{\begin{adjustbox}{max width=\textwidth}
\begin{tabular}{lccccc}
\toprule
Outcome & \makecell{Single-suggestion support} & \makecell{Dual-suggestion support} & \makecell{Mean difference (95\% CI)} & $P$ value & \makecell{Holm $P$} \\
\midrule
\multicolumn{6}{l}{\textit{GPT + Kimi: Group B vs Group A (11 cases per participant)}} \\
AI-assisted accuracy (\%) & $10.0 \pm 9.4$ & $8.4 \pm 10.7$ & $-1.56$ ($-5.98$ to 2.86) & .485 & .970 \\
Accuracy change (pp) & $-10.7 \pm 11.6$ & $-12.1 \pm 10.2$ & $-1.44$ ($-6.25$ to 3.37) & .553 & .553 \\
Confidence change (points) & $0.11 \pm 0.34$ & $0.11 \pm 0.24$ & 0.00 ($-0.13$ to 0.13) & .956 & 1.000 \\
\makecell[l]{Incorrect final diagnoses with\\increased confidence (\%)} & $20.5 \pm 22.1$ & $17.3 \pm 17.5$ & $-3.14$ ($-11.92$ to 5.64) & .479 & .957 \\
\addlinespace
\multicolumn{6}{l}{\textit{GPT + Gemini: Group C vs Group A (10 cases per participant)}} \\
AI-assisted accuracy (\%) & $5.3 \pm 9.1$ & $4.6 \pm 10.7$ & $-0.62$ ($-5.01$ to 3.78) & .781 & .970 \\
Accuracy change (pp) & $-15.3 \pm 13.6$ & $-19.8 \pm 14.1$ & $-4.51$ ($-10.62$ to 1.61) & .146 & .292 \\
Confidence change (points) & $0.09 \pm 0.32$ & $0.09 \pm 0.33$ & $-0.01$ ($-0.15$ to 0.14) & .921 & 1.000 \\
\makecell[l]{Incorrect final diagnoses with\\increased confidence (\%)} & $21.8 \pm 21.0$ & $19.3 \pm 22.4$ & $-2.48$ ($-12.08$ to 7.12) & .608 & .957 \\
\bottomrule
\end{tabular}
\end{adjustbox}}
\end{threeparttable}
\fulltablenote{Note.---Data are participant-level mean $\pm$ SD. Mean differences are dual-suggestion minus single-suggestion support.\par Accuracy change was calculated as AI-assisted minus AI-unassisted accuracy. Negative values indicate a smaller accuracy gain or a larger decrease with dual-suggestion support.\par Incorrect final diagnoses with increased confidence were final incorrect decisions for which confidence increased after AI assistance, expressed as a percentage of cases in the corresponding stratum.\par Comparisons used two-sided Welch $t$ tests. Holm adjustment covered the two prespecified dual-versus-single contrasts within each outcome.\par CI = confidence interval; pp = percentage points.}
\end{table}

\end{document}